\documentclass[10pt,journal,compsoc]{IEEEtran}
\ifCLASSOPTIONcompsoc
  \usepackage[nocompress]{cite}
\else
  \usepackage{cite}
\fi
\ifCLASSINFOpdf
\else
  Online Image Vectorizern (dvipsone, dvipdf, if not using dvips). graphicx
\fi

\usepackage{subfigure}
\usepackage{times}
\usepackage{epsfig}
\usepackage{hyperref}
\usepackage{graphicx}
\usepackage{amsmath}
\usepackage{amssymb}
\usepackage{multirow}
\usepackage{tabularx}
\usepackage{booktabs}

\usepackage{amsmath, bm}
\usepackage{algorithm,algpseudocode}
\usepackage{xcolor,colortbl}
\usepackage{color, colortbl}
\usepackage{breakcites}
\usepackage{dsfont}
\usepackage{bbm}
\usepackage{makecell}
\usepackage{colortbl}
\usepackage{marvosym}

\usepackage{pifont}

\newcommand{\ie}{\textit{i.e. }}
\newcommand{\eg}{\textit{e.g. }}

\newcommand{\etc}{\textit{etc}.}

\definecolor{myblue}{RGB}{23,183,241}
\definecolor{mygray}{RGB}{230,230,230}

\begin{document}

\title{Lowis3D: Language-Driven Open-World Instance-Level 3D Scene Understanding}

\author{Runyu Ding,~ Jihan Yang,~ Chuhui Xue,~ Wenqing Zhang,~ Song Bai,~ Xiaojuan Qi
    \IEEEcompsocitemizethanks{
        \IEEEcompsocthanksitem Runyu Ding, Jihan Yang and Xiaojuan Qi are with the Department of Electrical and Electronic Engineering at The University of Hong Kong, Hong Kong. Chuhui Xue, Wenqing Zhang and Song Bai are with ByteDance Inc. 
    
        \IEEEcompsocthanksitem Email: ryding@eee.hku.hk, jhyang@eee.hku.hk, xuec0003@e.ntu.edu.sg, wenqingzhang@bytedance.com, songbai.site@gmail.com, xjqi@eee.hku.hk
    
        \IEEEcompsocthanksitem Runyu Ding: Part of the work done during an internship at ByteDance Inc.
    }
}

\IEEEtitleabstractindextext{%
\begin{abstract}
   
 Open-world instance-level scene understanding aims to locate and recognize unseen object categories that are not present in the annotated dataset. 
 This task is challenging because the model needs to both localize novel 3D objects and infer their semantic categories. 
 A key factor for the recent progress in 2D open-world perception is the availability of large-scale image-text pairs from the Internet, which cover a wide range of vocabulary concepts. However, this success is hard to replicate in 3D scenarios due to the scarcity of 3D-text pairs. To address this challenge, we propose to harness pre-trained vision-language (VL) foundation models that encode extensive knowledge from image-text pairs to generate captions for multi-view images of 3D scenes. This allows us to establish explicit associations between 3D shapes and semantic-rich captions. Moreover, to enhance the fine-grained visual-semantic representation learning from captions for object-level categorization, we design hierarchical point-caption association methods to learn semantic-aware embeddings that exploit the 3D geometry between 3D points and multi-view images. In addition, to tackle the localization challenge for novel classes in the open-world setting, we develop debiased instance localization, which involves training object grouping modules on unlabeled data using instance-level pseudo supervision. This significantly improves the generalization capabilities of instance grouping and thus the ability to accurately locate novel objects. We conduct extensive experiments on 3D semantic, instance, and panoptic segmentation tasks, covering indoor and outdoor scenes across three datasets. Our method outperforms baseline methods by a significant margin in semantic segmentation (\eg 34.5\%$\sim$65.3\%), instance segmentation (\eg 21.8\%$\sim$54.0\%) and panoptic segmentation (\eg 14.7\%$\sim$43.3\%).
 Code will be available.
\end{abstract}

\begin{IEEEkeywords}
    3D scene understanding, instance segmentation, panoptic segmentation, point clouds, open vocabulary, open world.
\end{IEEEkeywords}}

\maketitle

\IEEEdisplaynontitleabstractindextext

\IEEEpeerreviewmaketitle

\IEEEraisesectionheading{\section{Introduction}\label{sec:inrto}}

\begin{figure*}[t]
    \begin{center}
    \includegraphics[width=1\linewidth]{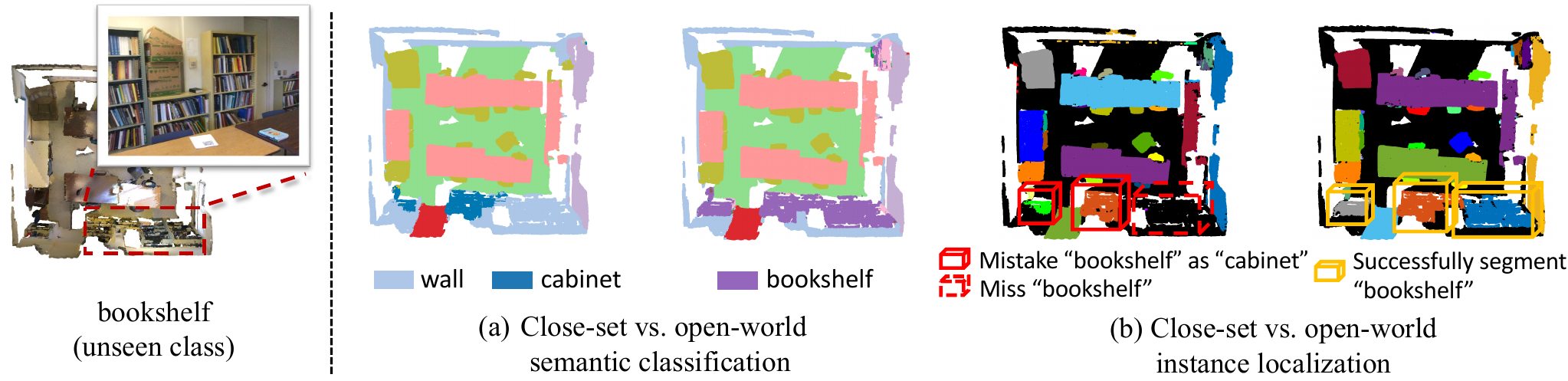}
    \end{center}
    \caption{An example of 3D open-world instance-level scene understanding on ScanNet~\cite{dai2017scannet}, where the unseen class is ``bookshelf''. 
    In this case, the close-set model mistakenly classifies the``bookshelf'' as a``cabinet'' or fails to recognize it entirely. However, our open-world model accurately localizes and recognizes the ``bookshelf''.}
    \label{fig:teaser}
\end{figure*}

\IEEEPARstart{3}{D} instance-level scene understanding,  which involves localizing 3D objects and understanding their semantics, is a crucial perception component for real-world applications such as virtual reality (VR), robot manipulation, and human-machine interaction. Deep learning has achieved remarkable success in this area~\cite{graham20183d,vu2022softgroup,misra2021-3detr}. However, deep models trained on human-annotated datasets can only comprehend semantic categories that are present in the dataset; that is, they are confined to close-set prediction. Consequently, they fail to recognize novel categories that are not seen in the training data, as shown in Fig.~\ref{fig:teaser}. This severely limits their applicability in real-world scenarios such as robotics and autonomous driving with unlimited potential categories. Furthermore, the high annotation costs on 3D datasets (\eg~22.3 minutes for a single scene with 20 classes~\cite{dai2017scannet}) make it impractical to rely solely on human labor to cover all real-world categories.
This motivates us to investigate open-world 3D instance-level scene understanding, which allows a model to recognize and localize open-set classes that are not included in the label space of an annotated dataset (see Fig.~\ref{fig:teaser}). This involves two key components: open-world semantic comprehension and open-world instance localization.

Recently, vision-language (VL) foundation models~\cite{radford2021learning,jia2021scaling,florence} have demonstrated the ability to learn effective vision-language embeddings that connects textual descriptions and corresponding images by training on web-crawled image data along with semantic-rich captions~\cite{sharma2018conceptual}. 
These embeddings are further leveraged to solve various 2D open-world tasks including object detection~\cite{gu2021open,Hanoona2022Bridging}, semantic segmentation~\cite{xu2021simple,li2022languagedriven,zhou2022maskclip}, panoptic segmentation~\cite{xu2023open} and \etc. 
Although the pre-training paradigm has significantly advanced open-vocabulary image understanding tasks, its direct applicability in the 3D domain is hindered by the lack of large-scale 3D-text pairs.

To address this challenge, some recent efforts~\cite{zhang2022pointclip,huang2022clip2point} have tried to convert the 3D data into 2D modalities such as RGB images and depth maps. By leveraging pre-trained VL foundation models, these methods aim to analyze the projected 2D data to enable open-world recognition of 3D objects. However, this line of methods has several major drawbacks, rendering it suboptimal for scene-level understanding such as instance segmentation. 
First, to represent a 3D scene, multiple RGB images and depth maps are needed for processing, which results in high memory and computation costs during both training and inference. Secondly, the projection from 3D to 2D causes information loss and prevents direct learning from geometry-rich 3D data, resulting in poor performance. Our preliminary study reveals the state-of-the-art 2D open-world semantic segmentation approach, MaskCLIP \cite{zhou2022maskclip}, can only achieve 17.8\% mIoU yet with a 20-fold increase in latency when tasked to segment projected 2D images from the 3D ScanNet dataset~\cite{dai2017scannet}.

Thus, inspired by the remarkable success of vision-language foundation models for various VL tasks~\cite{gu2021open,Hanoona2022Bridging,xu2021simple,li2022languagedriven,zhou2022maskclip,zhang2022pointclip,huang2022clip2point}, we ask: \textit{can we leverage the abundant knowledge encoded in VL foundation models to build an explicit association between 3D and language for open-world understanding?}
In pursuit of this goal, our core idea is to use pre-trained VL models~\cite{vit-gpt2,wang2022ofa} to caption readily-available image data that is aligned with 3D data –- specifically, the point set within the corresponding frustum that generates the image. These images can be obtained either through neural rendering~\cite{dai2020neural,yu2021pixelnerf} techniques or directly from the 3D scene collection pipeline~\cite{dai2017scannet}.In this way, we are able to transfer rich semantics to the 3D domain, thereby enabling an explicit connection between 3D data and vocabulary-rich text descriptions for open-world 3D scene understanding.

After establishing the point-language association, the subsequent question arises regarding how to empower a 3D network to acquire semantic-aware embeddings from (pseudo) captions. The primary obstacle lies in the complex object compositions in the 3D scene-level data (see Fig.~\ref{fig:caption}), which makes it hard to link objects with their corresponding words within the caption.
This is different from object-centric image data that typically consists of a single centered object~\cite{radford2021learning}. However, there is a fortunate aspect to consider: the 3D geometry relation between captioned multi-view images and a 3D scene can be exploited to construct hierarchical point-caption pairs. These pairs encompass captions at various levels, including scene-level, view-level, and entity-level captions, which provide coarse-to-fine supervision signals to enable the effective learning of visual-semantic representations from a rich vocabulary corpus through contrastive learning.

Although point-language association gives the model the strong ability to recognize novel semantic concepts, the model still struggles to correctly localize the 3D objects, leading to predictions of incomplete instance masks  or incorrectly predicting multiple instances as one (see PLA results in Fig.~\ref{fig:vis_2}).
This is because the existing close-set 3D instance localization network  tends to overfit annotated/base categories and thus easily fails to localize unseen objects with novel shapes, scales, or contexts. To the best of our knowledge, this problem has not been addressed in current open-world 3D scene understanding studies~\cite{Peng2023OpenScene,chen2023clip2scene}. To tackle this challenge, we propose a debiased instance localization module that provides instance-level pseudo supervision for clustering potential novel objects into candidate proposals. This module improves the localization ability of our framework for unseen objects, thereby rendering our method more effective for 3D open-world instance and panoptic segmentation tasks.

Overall, our holistic framework, named Lowis3D, combines point-language association for semantic recognition and debiased instance localization for object localization, offering a flexible and general solution for open-world 3D scene understanding. By comprehensively addressing the two essential problems of scene understanding, our framework provides a solid foundation for advancing the field of open-world 3D scene understanding.

We conduct extensive experiments on three scene understanding tasks across three popular large-scale datasets~\cite{dai2017scannet,armeni20163d,fong2022panoptic} covering both indoor and outdoor scenarios. Results show that Lowis3D significantly surpasses the baseline models, achieving improvements of 21.8\% $\sim$ 54.0\% hAP$_{50}$ on instance segmentation, 14.7\% $\sim$ 43.3\% hPQ on panoptic segmentation and 34.5\% $\sim$ 65.3\% hIoU on semantic segmentation,  manifesting its effectiveness. Besides, when compared with  PLA~\cite{ding2022language}, Lowis3D exhibits a performance gain of 2.4\% $\sim$ 12.6\% on  tasks that require instance-level understanding.
In addition, our model shows its scalability and extensibility by achieving 0.3\%$\sim$3.5\% improvements in semantic recognition when utilizing more advanced image-captioning model that provides higher-quality caption supervision. This further highlights the potential of our approach to adapt and excel with more advanced techniques.

\vspace{0.02in}
\noindent\textbf{Difference to our conference paper:} This manuscript substantially extends the conference version~\cite{ding2022language} in the following aspects. (i). We provide an in-depth analysis of the challenges in open-world 3D scene understanding in terms of unseen semantic recognition and instance localization, which helps to better understand and address this task. (ii). We propose a lightweight proposal grouping module that effectively reduces the bias toward base classes by incorporating pseudo-offset supervision signals. This greatly enhances the adaptability of instance localization for novel classes. (iii). We conduct extensive experiments on three large-scale scene understanding datasets that cover both indoor  and outdoor scenarios, surpassing PLA in instance-level understanding by a large margin. (iv.) We further attempt our Lowis3D on the 3D panoptic segmentation task, achieving significant improvements on nuScenes~\cite{fong2022panoptic} dataset. Overall, these enhancements contribute to a more comprehensive and effective framework for open-world 3D scene understanding with high potential and applicability in various real-world scenarios.


\section{Related Work}

\noindent\textbf{3D scene understanding} targets at comprehending the semantic meaning of objects and their surrounding environment through the analysis of point clouds. 
In this study, we  focus on three integral scene understanding tasks: semantic, instance and panoptic segmentation. 
\textit{3D semantic segmentation} aims to produce point-wise semantic predictions for point clouds. Representative works involves point-based architecture \cite{qi2017pointnet++,huang2018recurrent} with elaborately crafted point convolution operations~\cite{thomas2019KPConv,xu2021paconv}, transformers~\cite{lai2022stratified} that capture long-range point contexts with attention mechanisms, and voxel-based ~\cite{graham20183d,choy20194d} approaches using efficient 3D sparse convolutions \cite{graham2017submanifold} to generate context-aware predictions. 
\textit{3D instance segmentation} goes a step further by distinguishing distinct object instances based on semantic segmentation. 
Existing methods typically adopt either a top-down solution~\cite{yi2019gspn,yang2019learning}, that is to predict the 3D bounding box followed by the mask refinement, or a bottom-up~\cite{jiang2020pointgroup,vu2022softgroup} approach through predicting point offsets towards object centers and grouping points into mask proposals. \textit{3D panoptic segmentation}, on the other hand, strives to unify instance and semantic predictions to generate coherent scene segmentation. Based on how to obtain instance IDs, it can be coarsely categorized into proposal-based stream~\cite{hurtado2020mopt} with top-down proposal generation manners and proposal-free stream~\cite{milioto2020lidar,hong2021lidar} with bottom-up instance grouping approaches. Though achieving promising results on close-set benchmarks, existing methods struggle to recognize or localize open-set novel categories. Addressing this limitation is the main focus of our work.

\vspace{0.05in}\noindent\textbf{Open-world learning} targets at recognizing novel classes that are not present in training annotations. Early approaches primarily adhere to the zero-shot setting, which can be coarsely categorized into generative methods~\cite{bucher2019zero,gu2020context} and discriminative methods~\cite{xian2019semantic,baek2021exploiting}. 3DGenZ~\cite{michele2021generative} extends~\cite{bucher2019zero} to the realm of 3D understanding for zero-shot semantic segmentation.  
Moving beyond the zero-shot learning, the more general open-world setting presumes the accessibility of a large vocabulary bank during the training phase~\cite{zareian2021open}. In the context of \textit{2D open-world learning}, existing approaches take different approaches. Some leverage massive annotated image-caption pairs to provide weak supervision for vocabulary enhancement~\cite{zareian2021open,zhou2022detecting}. Others utilize pre-trained vision-language (VL) models, such as CLIP~\cite{radford2021learning} that is trained on extensive image-caption pairs to tackle open-world understanding. 

In comparison, \textit{3D open-world learning} is still in its infancy with only a few endeavors so far. Some papers~\cite{zhang2022pointclip,huang2022clip2point} focus on object-level classification. They explore techniques to project object-level 3D point clouds onto multi-view 2D images and depth maps, and leverage the pre-trained VL model for producing open-world predictions.
Nevertheless,they suffer from heavy computation and subpar performance when applied to 3D scene understanding tasks.
More recent work~\cite{Peng2023OpenScene,jatavallabhula2023conceptfusion,yang2023regionplc} address semantic-level scene understanding by aligning 3D points with 2D boxes or pixels and distilling dense semantic-aware embeddings, which relies on time-consuming image processing or heavy disk storage. 
In this work, we focus on instance-level scene understanding, proposing a language-driven 3D open-world paradigm that learns visual-semantic embeddings and a debised instance localization for generalizable objectness learning.
Our Lowis3D framework can be generally applied to various scene understanding tasks and offers efficiency with only the 3D network deployed in training and inference.

\section{Preliminary}\label{sec:method_pre}

\begin{figure*}[t]
    \begin{center}
    \includegraphics[width=1\linewidth]{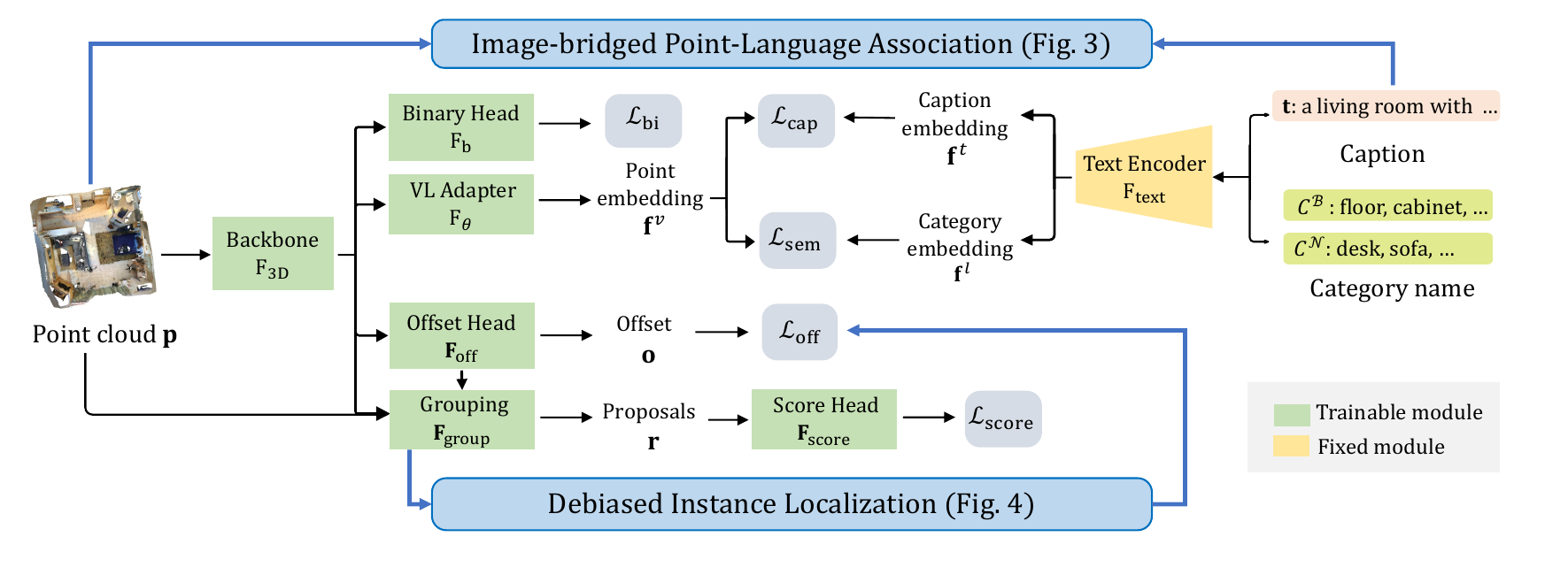}
    \end{center}
    \caption{Our language-driven 3D instance-level scene understanding framework that can handle open-world queries. The model learns rich semantics through point embeddings that are aligned with caption embeddings using point-language association (details in Fig.~\ref{fig:caption}). A binary head is used to adjust predicted semantic scores based on the probabilities of belonging to base and novel classes. A debiased instance localization module generates confident pseudo supervisions on novel categories to enhance the open-world objectness learning (details in Fig.~\ref{fig:inst_group}).
    Best viewed in color.}
    \label{fig:framework}
\end{figure*}

3D open-world instance-level segmentation targets at localizing and recognizing unseen categories without using human annotation as supervision. 
Formally, annotations on semantic and instance levels $\mathcal{Y} = \{(\mathbf{y}_{\text{sem}}, \mathbf{y}_{\text{ins}})\}$ are divided into two sets: base categories $\mathcal{C}^B$ and novel categories $\mathcal{C}^N$. During the training phase, the 3D model has access to all point clouds $\mathcal{P}=\{\textbf{p}\}$, but it only has annotations for the base classes, denoted as $\mathcal{Y}^B$. The model is unaware of the annotations $\mathcal{Y}^N$ and the category names associated with the novel classes $\mathcal{C}^N$.
However, the 3D model is required to localize objects and classify points belonging to both the base and novel categories $\mathcal{C}^B \cup \mathcal{C}^N$ during inference.

A typical 3D instance understanding network consists of a 3D encoder $\text{F}_{\text{3D}}$ for feature extraction, a dense classification head $\text{F}_{\text{sem}}$ for semantic comprehension, and an instance head for instance localization and mask prediction. Specifically, we use a bottom-up strategy for the instance head that includes an offset branch $\text{F}_{\text{off}}$ to predict point offsets towards object centers, an instance grouping module $\text{F}_{\text{group}}$ to cluster offset-shifted points into proposals, and a proposal scoring network $\text{F}_{\text{score}}$ to score each proposal for post-processing and confidence ranking. The  inference pipeline is shown below:
\begin{gather}
    \mathbf{f}^p=\text{F}_{\text{3D}}(\mathbf{p}), ~~
    \mathbf{s}= \sigma \circ \text{F}_{\text{sem}}(\mathbf{f}^p),\label{eq:sem_basic}\\
    \mathbf{o}=\text{F}_{\text{off}}(\mathbf{f}^p), ~~
    \mathbf{r}=\text{F}_{\text{group}}(\mathbf{p}, \mathbf{o}, \mathbf{s}), ~~ \mathbf{z}=\text{F}_{\text{score}}(\mathbf{r}, \mathbf{f}^p),\label{eq:inst_basic}
\end{gather}
where $\mathbf{p}$ is the input point cloud, $\mathbf{f}^p$ is the point-wise 3D feature,  $\sigma$ is the softmax function, $\mathbf{s}$ is the semantic score, $\mathbf{o}$ is the point offset,  $\mathbf{r}$ is the grouped proposal, and $\mathbf{z}$ is the proposal scores. With these network predictions, we can then calculate semantic classification loss $\mathcal{L}_\text{sem}$ with semantic label $\mathbf{y}_\text{sem}$, point offset loss $\mathcal{L}_\text{off}$ with offset label $\mathbf{y}_\text{offset}$ as well as proposal scoring loss $\mathcal{L}_\text{score}$ with proposal label $\mathbf{y}_\text{ppl}$ similar to~\cite{jiang2020pointgroup,vu2022softgroup} as Eq.~\eqref{eq:basic_loss} and Eq.~\eqref{eq:basic_loss_2}, where the $\mathbf{y}_\text{offset}$ and $\mathbf{y}_\text{ppl}$ can be obtained from $\mathbf{y}_\text{ins}$. Notice that during training $\mathbf{y}_\text{sem}$ and $\mathbf{y}_\text{ins}$ only relate to base categories $\mathcal{C}^B$.
\begin{gather}\label{eq:basic_loss}
    \mathcal{L}_{\text{sem}}=\text{Loss}(\mathbf{s}, \mathbf{y}_\text{sem}), \\
    \mathcal{L}_{\text{off}}=\text{Loss}(\mathbf{o}, \mathbf{y}_\text{off}), ~~ \mathcal{L}_{\text{score}}=\text{Loss}(\mathbf{z}, \mathbf{y}_\text{ppl})\label{eq:basic_loss_2}.
\end{gather}
For panoptic segmentation, we fuse semantic prediction $\mathbf{s}$ with instance proposals $\mathbf{r}$ to generate a coherent segmentation map following~\cite{fong2022panoptic}.

\section{Open-World Instance-level Scene Understanding and Challenges}\label{sec:analysis}

This section elaborates on our design to extend the close-set network into an open-world leaner. We then analyze its main challenges to achieve optimal performance on open-world tasks.

\subsection{Open-World Setups}
Although it is possible to train a scene understanding model using the loss functions in Eq.~\eqref{eq:basic_loss}, the resulting model is actually a close-set model with a close-set classifier $\text{F}_{\text{sem}}$ and a close-set design in proposal grouping generation using $\text{F}_{\text{off}}$, $\text{F}_{\text{group}}$, and $\text{F}_{\text{score}}$. As a close-set model, it is unable to handle the task of recognizing or localizing unseen categories.
To address this issue,a text-embedded semantic classifier is introduced to obtain an open-world model.
Furthermore, we modify the instance prediction branch into a class-agnostic one that can be naturally extended to arbitrary categories.

\subsubsection{Text-Embedded Semantic Classifier}\label{sec:text_embed_cls}
First, as shown in Fig.~\ref{fig:framework}, to enable the model to become an open-world learner, we 
replace its learnable semantic classifier $\text{F}_\text{sem}$ with pre-trained category text embeddings $\mathbf{f}^l$ and a learnable vision-language adapter $\text{F}_\theta$ to align the dimension between 3D features $\mathbf{f}^p$ and $\mathbf{f}^l$ as follows, 
\begin{equation}\label{eq:text_encoder}
    \mathbf{f}^v=\text{F}_\theta(\mathbf{f}^p), ~~ \mathbf{s}= \sigma(\mathbf{f}^l\cdot \mathbf{f}^v),
\end{equation}
where $\mathbf{f}^v$ is the projected feature obtained through the VL adapter $\text{F}_\theta$, $\mathbf{f}^l = [\mathbf{f}^l_1, \mathbf{f}^l_2, \cdots, \mathbf{f}^l_{k}]$ is the category embeddings generated by encoding $k$ category names $\mathcal{C}$ with a frozen text encoder $\text{F}_{\text{text}}$ such as CLIP~\cite{radford2021learning} BERT~\cite{devlin2018bert} (see Fig.~\ref{fig:framework}). To make predictions, the model computes the cosine similarity between the projected point embeddings $\mathbf{f}^v$ and the 
category embeddings $\mathbf{f}^l$ and and then selects the category with the highest similarity as the prediction. During training,  the embeddings $\mathbf{f}^l$ only include those belonging to base classes $\mathcal{C}^B$. However, during open-world inference, the embeddings related to both base and novel categories $\mathcal{C}^B \cup \mathcal{C}^N$ are utilized.
By employing the category embeddings $\mathbf{f}^l$ as a classifier, the model gains the capability to perform open-world inference on any desired categories. We name this design as \textbf{OV-SparseConvNet} as a semantic baseline.

\subsubsection{Semantic-Guided Instance Module}\label{sec:instance_module}
Basically, we adopt the instance head from SoftGroup~\cite{vu2022softgroup} for instance segmentation, as shown in Fig.~\ref{fig:framework}.
The offset head $\mathbf{F}_\text{off}$ predicts class-agnostic offsets $\mathbf{o}$ for each point towards the object center. During training, only proposals belonging to base classes receive supervisions and undergo grouping. However, we can perform grouping for any novel categories during open-world inference due to the open-vocabulary capabilities of the semantic scores $\mathbf{s}$ obtained through the text-embedded classifier. Additionally, we do not use class statistics (\ie the average number of points per instance mask for each class) to assist grouping here since they are not available for novel categories.

For the proposal scoring head $\textbf{F}_\text{score}$, to facilitate its adaptability to novel categories, we make modifications to its functionality. Specifically, it now outputs class-agnostic binary scores, serving as indicators of the objectness for each proposal, instead of producing per-class confidence scores. This modification eliminates inherent biases towards seen categories and enables better generalization to novel categories. Additionally, this also allows us to train the proposal scoring network without prior knowledge of the novel categories that lie beyond the existing vocabulary. Furthermore, we remove the proposal classification head designed in SoftGroup to avoid overfitting to base categories and choose to aggregate semantic scores $\textbf{s}$ from our text-embedded $\textbf{F}_\text{sem}$ for each proposal. Since $\textbf{F}_\text{sem}$ owns strong open-vocabulary capabilities, we can use it to predict arbitrary novel categories. We call this baseline model \textbf{OV-SoftGroup}, which can perform open-vocabulary instance and panoptic segmentation.

\subsection{Challenges}
With a text-embedded classifier and a class-agnostic instance grouping module, we obtain a deep model that can perform open-world instance-level scene understanding. However, our experiments show that this model suffers from poor generalization to novel categories after training only on base classes. Therefore, we investigate the difficulties in 3D open-world instance-level scene understanding and identify the key challenges related to semantic recognition and instance localization.

\subsubsection{Challenges on Semantic Understanding}\label{sec:challenge_sem}
We first train OV-sparseConvNet on $\mathcal{C}^B$ and evaluate its performance on $\mathcal{C}^B \cup \mathcal{C}^N$. Table~\ref{tab:sem_analysis} shows that the model fails to recognize novel classes in the ScanNet dataset, with a large mIoU gap of about 79\% compared to the fully-supervised model (the model is trained on $\mathcal{C}^B \cup \mathcal{C}^N$). We empirically identify two factors contributing to this substantial gap: the model's bias towards the base categories and its inability to comprehend the semantic meaning of unseen categories.

Firstly, we observe that the model performs poorly on novel categories, achieving zero mIoU. Moreover, it exhibits an approximate 34\% performance gap when compared to OV-sparseConvNet$^\dagger$, which infers points from base and novel classes separately to avoid confusion between the two category splits. It demonstrates that the model often misclassifies novel categories as base ones, indicating a \textit{strong bias toward base categories}.

We then investigate the performance of OV-SparseConvNet$^\dagger$. Even without the influence of overfitting to base categories, it still performs poorly, with an about 45\% mIoU gap on novel categories compared to a fully-supervised model. Such a performance gap can be attributed to the model's inability to distinguish different novel categories, indicating a lack of understanding of unseen categories and poor generalization to novel concepts.

\begin{table}[htbp]
 \caption{Investigation of the semantic performance gap between OV-SparseConvNet and fully-supervised model on ScanNet with 15 base categories and four novel categories in terms of mIoU. $\dagger$ denotes forcing semantic predictions to fit the correct partition, \ie $C^\mathcal{B}$ or $C^\mathcal{N}$.}
    \label{tab:sem_analysis}
    \centering
    \setlength\tabcolsep{6pt}
    \scalebox{1.0}{
        \begin{tabular}{l|cc}
            \bottomrule[1pt]
            \multirow{2}{*}{Method} & \multicolumn{2}{c}{ScanNet} \\
            \cline{2-3}
            & base mIoU & novel mIoU  \\
            \hline
            OV-SparseConvNet & 64.4 & 00.0 \\
            OV-SparseConvNet$^\dagger$ & 70.7 & 34.3  \\
            Fully-Sup. & 68.4 & 79.1 \\
            \toprule[0.8pt]
        \end{tabular}
    }
\end{table}

\subsubsection{Challenges on Instance Localization} \label{sec:challenge_ins}
Similarly, we investigate the open-world instance localization ability with OV-SoftGroup.
We train OV-SoftGroup on $\mathcal{C}^B$ and evaluate its performance on $\mathcal{C}^B \cup \mathcal{C}^N$. We use the point offset error (MAE) to assess the offset head $\mathbf{F}_\text{off}$ and the average recall (AR) to measure the quality of grouped instance proposals.
Table~\ref{tab:inst_analysis} reveals that our OV-SoftGroup, despite its class-agnostic $\mathbf{F}_\text{off}$ for instance grouping, experiences \textit{overfitting to object patterns of base categories}, with the larger offset error compared to fully-supervised model. Additionally, the lower AR reflects the poorer quality of proposals, further confirming this issue. 
This is potentially due to the fact that unseen objects may have novel shapes, sizes, and contexts that differ from base categories, which makes the knowledge learned from base categories not generalizable to novel ones. This challenge is often ignored in existing open-world studies~\cite{Peng2023OpenScene,chen2023clip2scene}, which we will tackle in this paper.

\begin{table}[htbp]
    \caption{Investigation of instance performance gap between OV-SoftGroup and fully-supervised model on ScanNet  with 13 base categories and 4 novel categories in terms of AR$_{50}$ (average recall at IoU threshold 0.5) and offset mAE (mean absolute error).}
    \label{tab:inst_analysis}
    \centering
    \setlength\tabcolsep{2pt}
    \scalebox{1.0}{
        \begin{tabular}{l|cc|cc}
            \bottomrule[1pt]
            \multirow{2}{*}{Method} & \multicolumn{4}{c}{ScanNet} \\
            \cline{2-5}
             & base mAE ($\downarrow$) & novel mAE ($\downarrow$) & base AR ($\uparrow$)& novel AR ($\uparrow$) \\
            \hline
            OV-SoftGroup & 0.37 & 0.68 & 47.2 & 21.7 \\
            Fully-Sup. & 0.36 & 0.46 & 47.3 & 57.0 \\
            \toprule[0.8pt]
        \end{tabular}
    }
\end{table}

\section{Method}

To address the challenges discussed in Section~\ref{sec:analysis}, we propose a holistic pipeline for open-world 3D instance-level scene understanding called Lowis3D. Our framework consists of a point-language association module (see Sec.~\ref{sec:image-language}) that leverages the powerful VL foundation models for learning visual-semantic relationships. This helps expose the model to novel concepts beyond the annotated dataset without requiring human annotations. Besides, we introduce a binary prediction head for distinguishing novel and base categories for calibrating biased predictions among base and novel categories (see Sec.~\ref{sec:binary}). 
Finally, we design debiased instance localization to enhance objectness learning and facilitate object grouping on novel categories (see Sec.~\ref{sec:debiased_inst}).

\begin{figure*}[t]
    \begin{center}
    \includegraphics[width=1\linewidth]{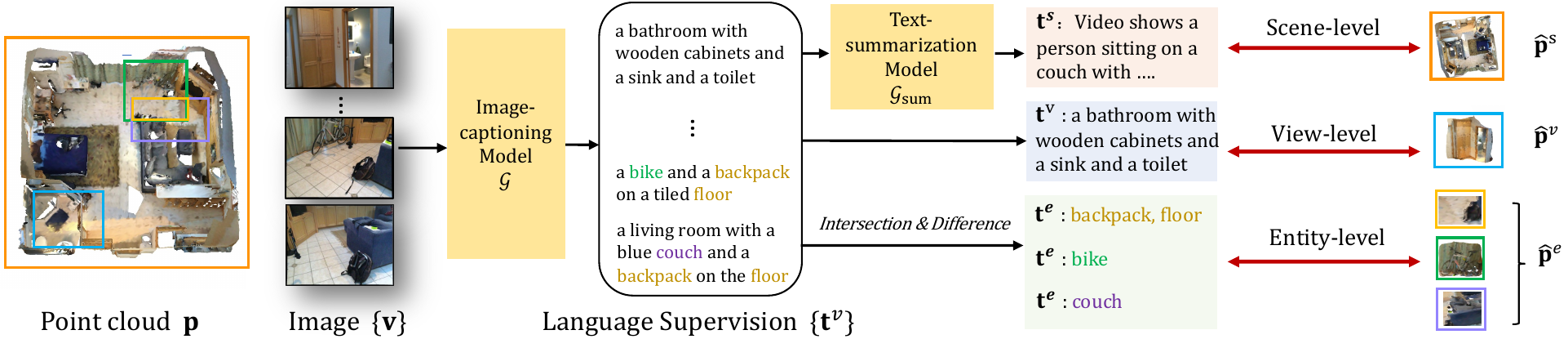}
    \end{center}
    \caption{Image-bridged point-language association. We present hierarchical point-language association manners at scene-level, view-level and entity-level, which assign coarse-to-fine point sets with caption supervision through vision-language foundation models and multi-view RGB images.}
    \label{fig:caption}
\end{figure*}

\subsection{Image-Bridged Point-Language Association}\label{sec:image-language}

As shown in Table~\ref{tab:sem_analysis}, OV-SparseConvNet performs poorly on novel categories due to its limited semantic recognition capabilities. 
Recent open-vocabulary works~\cite{li2022languagedriven,Hanoona2022Bridging,gu2021open} in the 2D domain have shown the effectiveness of using language supervision to train vision backbones on large-scale text-image paired data. The large-scale vision-language dataset provides rich language supervision that enables the vision backbone to access a wide range of semantic concepts with a large vocabulary and helps to align vision and language features. This enhances the generalization of novel concepts. However, this success is hard to achieve in 3D due to the lack of Internet-scale paired 3D-text data.

To tackle this challenge, we propose an image-bridged point-language association module that provides language supervision for 3D scene understanding without the need for human annotation, as illustrated in  Fig.~\ref{fig:framework} and Fig.~\ref{fig:caption}. 
Our core idea is to leverage multi-view images from a 3D scene as a bridge to access the knowledge encoded in vision-language foundation models for generating language descriptions. 
As shown in Fig.~\ref{fig:caption}, an image of a 3D scene is input to a powerful image-captioning model, which generates a text description. Then, the text description is associated with a point set in the 3D scene utilizing the geometric correspondence between the image and the 3D scene. In the following, we provide more details about our captioning procedure and the designed hierarchical point-caption association.

\subsubsection{Multi-View Images Captioning}\label{sec:image-language}
With the development of multimodal vision and language learning, many foundation models~\cite{wang2022ofa,vit-gpt2,mokady2021clipcap} trained with extensive image-text pairs are readily available to solve the image captioning task~\cite{hossain2019comprehensive}.
Given the $j^\text{th}$ image of the $i^\text{th}$ scene, a pre-trained image-captioning model $\mathcal{G}$ can generate the corresponding text description $\textbf{t}^v_{ij}$:  
\begin{equation}
    \textbf{t}^v_{ij} = \mathcal{G}(\textbf{v}_{ij}). ~~
\end{equation}
Remarkably, despite $\mathcal{G}$ not being explicitly trained on a 3D scene understanding dataset such as ScanNet~\cite{dai2017scannet}, the generated captions are able to encapsulate the entire semantic label space of such datasets. Additionally, the captions $\textbf{t}$ provides fairly precise and comprehensive descriptions of various aspects, including room types, semantic categories with texture and color attributes, as well as spatial relationships. This is evident in the language supervision examples ${\mathbf{t}^v}$ shown in Fig.~\ref{fig:caption}, and additional examples can be found in the Appendix~\ref{sec:caption_exps}.

\subsubsection{Point Cloud Association with Language}\label{sec:3D-language}
After obtaining the image-text pairs, the subsequent step is to associate a point set $\mathbf{\hat{p}}$ with caption $\mathbf{t}$, using images $\textbf{v}$ as a bridge:
\begin{equation}
    \text{Explore}~\langle\mathbf{\hat{p}},\mathbf{t}\rangle ~\text{with}~\langle\mathbf{\hat{p}, \mathbf{v}}\rangle ~\text{and}~\langle\mathbf{v},\mathbf{t}\rangle.
\end{equation}
We propose three association fashions for point sets at varying spatial scales.

\vspace{0.05in}\noindent\textbf{Scene-Level Point-Language Association.} 
The coarsest and simplest association manner is to link caption supervision to all points within a specified 3D point cloud scene $\mathbf{\hat{p}}^s = \mathbf{p}$.  
As depicted in Fig.~\ref{fig:caption}, we consider all image captions ${\mathbf{t}_{ij}^v}$ associated with a given scene $\mathbf{p}_j$. These captions are used to generate a scene-level caption $\mathbf{t}^s_j$ by employing a text summarizer~\cite{lewis2019bart} $\mathcal{G}_\text{sum}$ as follows:
\begin{equation}
    \mathbf{t}^s_j=\mathcal{G}_\text{sum}(\{\mathbf{t}_{1j}^v,\mathbf{t}_{2j}^v,\cdots \mathbf{t}_{n_jj}^v\}),
\end{equation}
where $n_j$ is the total number of images for the scene $\mathbf{p}_j$.
By enabling each 3D scene $\mathbf{\hat{p}}^s$ to learn from its corresponding scene descriptions $\mathbf{t}^s$, we introduce a rich vocabulary and strengthen the visual-semantic relationships, enhancing the semantic understanding capability of the 3D backbone.
Despite the simple nature of scene-level language supervision, our empirical findings suggest that it can bolster the model's open-world ability by a significant margin (see Sec.~\ref{sec:ablation}). 

\vspace{0.05in}
\noindent\textbf{View-Level Point-Language Association.}
Albeit proven to be effective, scene-level language supervision assigns a single caption to all points in a scene, which neglects the relationship between the language and local 3D point clouds. Thus, it may not be optimal for instance-level scene understanding tasks.
To this end, we further propose a view-level point-language association manner that utilizes the geometrical relation between images and points to align each image caption $\mathbf{t}^v$ with a point set ${\mathbf{\hat{p}}}^v$ within the 3D view frustum of the corresponding image $\mathbf{v}$ (indicated by the \textcolor{myblue}{blue} box in Fig.~\ref{fig:caption}). 
Specifically, we obtain the view-level point set $\mathbf{\hat{p}^v}$ in the following steps. The RGB image is first back-projected $\mathbf{v}$ onto the 3D space with the assistance of the depth information $\mathbf{d}$ to get its corresponding point set $\mathbf{\ddot{p}}$: 
\begin{equation}\label{eq:projection}
    \left[\begin{array}{c|c}
        \mathbf{\ddot{p}} & \mathbf{1}
    \end{array}\right]
     = \mathbf{T}^{-1}
      \left[\begin{array}{c|c}
        \mathbf{v} & \mathbf{d}
    \end{array}\right],
\end{equation}
where $[\cdot|\cdot]$ denotes block matrix,  
$\mathbf{T} \in \mathbbm{R}^{3\times4}$ is the projection matrix derived from the camera intrinsic matrix and rigid transformations, typically obtained through sensor configurations or established SLAM approaches such as ~\cite{dai2017bundlefusion}.
Since the back-projected points $\mathbf{\ddot{p}}$ and points in 3D scene $\mathbf{p}$ may only have partial overlap, we then compute the overlapped regions between them to obtain the view-level point set $\mathbf{\hat{p}}^v$ as follows, 
\begin{equation}\label{eq:overlap}
    \mathbf{\hat{p}}^v = V^{-1}(R(V(\mathbf{\ddot{p}}), V(\mathbf{p}))),
\end{equation}
where $V$ and $V^{-1}$ denote the voxelization and reverse-voxelization processes, and $R$ means the radius-based nearest-neighbor search~\cite{Zhou2018}.
This view-based association approach enables the model to learn from region-level language descriptions, significantly augmenting the model's localization and recognition and capabilities for previously unseen categories.

\vspace{0.05in}\noindent\textbf{Entity-Level Point-Language Association.} 
While the view-level captioning strategy allows each image-caption pair $\mathbf{t}^v$ to be associated with a specific subset of the point cloud for a 3D scene, this association is still based on a large 3D area (\ie around 25K points) containing multiple semantic objects/categories, as illustrated in Fig.~\ref{fig:caption}. 
This broad coverage could be challenging for the 3D network to learn fine-grained point-wise semantic attributes and instance-aware position information from the language supervision. 
To this end, we further propose a fine-grained point-language association manner that owns the potential to construct entity-level point-caption pairs. 
In this way, each object instance is associated with a specific caption, allowing for more precise and detailed supervision.

Specifically, as depicted in Fig.~\ref{fig:caption}, we exploit the intersections and differences between adjacent view-level point sets $\mathbf{\hat{p}}^v$ and their corresponding view captions $\mathbf{t}^v$ to determine the associated points $\mathbf{\hat{p}}^e$ and caption $\mathbf{t}^e$ at entity level. 
To be specific, we first compute entity-level caption $\mathbf{t}^e$ as below:
\begin{gather}\label{eq:diff_and_intersect}
    w_i=E(\mathbf{t}_{i}^v), \\
    w^e_{i\setminus j}=w_i\setminus w_j,~~
    w^e_{j\setminus i}=w_j\setminus w_i,~~
    w^e_{i\cap j}=w_i\cap w_j,\\
    \mathbf{t}^e=\text{Concate}(w^e),
\end{gather}
where $E$ means the process of extracting a set of entity words $w$ from the caption $\mathbf{t}^v$, 
$\cap$ and $\setminus$ represent the set intersection and difference, respectively, and Concate means the concatenation of all words with spaces to form the entity-level caption $\mathbf{t}^e$.
Similarly, we can easily compute entity-level point sets and associate them with previously obtained entity-level captions to form point-caption pairs as follows:

\begin{gather}\label{eq:diff_and_intersect2}
    \mathbf{\hat{p}}^e_{i\setminus j} = (\mathbf{\hat{p}}^v_i\setminus \mathbf{\hat{p}}^v_j), ~~ \mathbf{\hat{p}}^e_{j\setminus i}=(\mathbf{\hat{p}}^v_j \setminus \mathbf{\hat{p}}^v_i),  ~~
    \mathbf{\hat{p}}^e_{i\cap j} = \mathbf{\hat{p}}^v_i \cap \mathbf{\hat{p}}^v_j,\\
    <\mathbf{\hat{p}}^e_{i\setminus j}, \mathbf{t}^e_{i\setminus j}>,<\mathbf{\hat{p}}^e_{j\setminus i}, \mathbf{t}^e_{j\setminus i}>, <\mathbf{\hat{p}}^e_{i\cap j}, \mathbf{t}^e_{i\cap j}>.
\end{gather}

After obtaining the entity-level $\langle\mathbf{\hat{p}}^e$, $\mathbf{t}^e\rangle$ pairs, we further apply filtering to ensure that each entity-level points set $\mathbf{\hat{p}}^e$ corresponds to at least one entity and is concentrated within a sufficiently small 3D space, as detailed below,

\begin{equation}\label{eq:filter}
    \gamma < |\mathbf{\hat{p}}^e| < \delta\cdot\min(|\mathbf{\hat{p}}^v_{i}|, |\mathbf{\hat{p}}^v_{j}|) ~~\text{and}~~ |\mathbf{t}^e| > 0,
\end{equation}
where $\gamma$ denotes a scalar to determine the minimal number of points, $\delta$ is a ratio that controls the maximal size of $\mathbf{\hat{p}}^e$, and the caption $\mathbf{t}^e$ must not be empty. This constraint assists in focusing on a fine-grained 3D point sets, thereby ensuring that there are fewer entities associated with each caption supervision.

\begin{table}[htbp]
    \centering
    \caption{Comparison among different point-language association manners.}
    \setlength\tabcolsep{3pt}
    \scalebox{1.0}{
        \begin{tabular}{l|ccc}
            \bottomrule[1pt]
             & scene-level & view-level & entity-level\\
             \hline
             \# points for each caption & 145,171 & 24,294 & 3,933\\
            \hline
            \# captions & 1,201 & 24,902 & 6,163 \\
            \hline
            complexity & simplest & middle & hardest \\
            \toprule[0.8pt]
        \end{tabular}
    }
    \label{tab:caption_statistics}
\end{table}

\vspace{0.05in}\noindent\textbf{Comparison among Different Point-Language Association Manners.}
The aforementioned three point-language association manners, arranged in a coarse-to-fine fashion, each possess different merits and limitations. 
As demonstrated in Table~\ref{tab:caption_statistics}, the scene-level association, while the simplest to implement, offers the coarsest correspondence between points and captions, with each caption corresponding to an average of over 140K points. 
On the other hand, the view-level association provides a finer level of point-language mapping, with a larger semantic space (over 20 times more captions) and a more localized point set (about 6 times fewer points for each caption) compared to the scene-level association.
The entity-level association establishes the most fine-grained correspondence relation, relating each caption with an average of only 4K points. This fine-grained association can contribute significantly to dense prediction and instance localization tasks. Our empirical results in Sec.~\ref{sec:ablation} demonstrate that the fine-grained association and a semantic-rich vocabulary space are two critical factors for open-world perception.

\subsubsection{Contrastive Point-Language Training}
~\label{sec:caption_formulation}
Having obtained point-caption pairs $\langle\mathbf{\hat{p}}, \mathbf{t}\rangle$, we can now guide the 3D backbone F$_\text{3D}$ to learn from semantic-rich language supervision. To achieve this, we introduce a general point-caption feature contrastive learning that can be applied to all types of coarse-to-fine point-language pairs.

First, we can obtain language embeddings $\mathbf{f}^t$ through a pre-trained text encoder $\text{F}_{\text{text}}$. 
Regarding the associated partial point set $\mathbf{\hat{p}}$, we 
select its corresponding point-wise features of adapted features $\mathbf{f}^v$ and employ the global average pooling to obtain its feature vector $\mathbf{f}^{\hat{p}}$ as follows, 

\begin{equation}\label{eq:caption_features}
    \mathbf{f}^t = \text{F}_{\text{text}}(\mathbf{t}), ~~ \mathbf{f}^{\hat{p}} = \text{Pool}(\mathbf{\hat{p}}, \mathbf{f}^v).
\end{equation}
Next, we apply the contrastive loss as~\cite{zareian2021open} to bring the corresponding point-language embeddings closer together and push away unrelated point-language embeddings. This loss function is defined as follows:

\begin{equation}\label{eq:contrastive}
    \mathcal{L}_{\text{cap}}=-\frac{1}{n_t}\sum\limits_{i=1}^{n_t} \log \frac{\exp(\mathbf{f}^{\hat{p}}_i\cdot\mathbf{f}^t_i/\tau)}{\sum_{j=1}^{n_t} \exp(\mathbf{f}^{\hat{p}}_i\cdot\mathbf{f}^t_j/\tau)},
\end{equation}
where $n_t$ represents the number of point-language pairs in any given association fashion and $\tau$ is a learnable temperature used to modulate the logits as CLIP~\cite{radford2021learning}. 
Additionally, to avoid noisy optimization and ensure effective learning, we remove duplicate captions within a batch during contrastive learning. Our final caption loss is a weighted combination of these losses and can be expressed as follows,

\begin{equation}\label{eq:cap}
\mathcal{L}^{\text{all}}_\text{cap}=\alpha_1*\mathcal{L}_\text{cap}^s+\alpha_2*\mathcal{L}_\text{cap}^v+\alpha_3*\mathcal{L}_\text{cap}^e,
\end{equation}
where $\alpha_1$, $\alpha_2$ and $\alpha_3$ are trade-off factors.

\subsection{Semantic Calibration with Binary Head}\label{sec:binary}
In Section~\ref{sec:challenge_sem}, we discussed the issue of over-confident semantic predictions on base classes and the calibration problem that arises as a result~\cite{guo2017calibration}. To address this issue, we propose a binary calibration module that rectifies semantic scores by considering the probability of a point belonging to either base or novel categories.

Specifically, as depicted in Fig.~\ref{fig:framework}, we employ a binary head $\text{F}\text{b}$ to distinguish between annotated (base) and unannotated (novel) points.
During training, $\text{F}_\text{b}$ is optimized with:
\begin{equation}
    \mathbf{s}_b = \text{F}_\text{b}(\mathbf{f}^p), ~~ \mathcal{L}_\text{bi}=\text{BCELoss}(\mathbf{s}_b, \mathbf{y}_b),
\end{equation}
where BCELoss($\cdot$, $\cdot$) denotes the binary cross-entropy loss, $\mathbf{y}_b$ denotes the binary label and $\mathbf{s}_b$ is the predicted binary score indicating the probability that a point belongs to novel categories. 
During the inference stage, the binary probability 
$\mathbf{s}^b$ is leveraged to correct the over-confident semantic score $\mathbf{s}$ as follows:
\begin{equation}\label{eq:binary}
    \mathbf{s} = \mathbf{s}^B\cdot (1-\mathbf{s}_b)+\mathbf{s}^N\cdot \mathbf{s}_b,
\end{equation}
where $\mathbf{s}^B$ is the semantic score calculated solely on base categories with novel class scores set to zero.
Similarly, $\mathbf{s}^N$ is computed only for novel classes, setting base class scores to zero. Notably, this calibration technique is also employed in instance and panoptic segmentation, specifically for calibrating the class predictions of grouped instance proposals. In Section~\ref{sec:ablation}, we provide empirical evidence to demonstrate that the probability calibration significantly improves the performance of both base and novel categories.
This demonstrates the effectiveness of our design in rectifying over-confident semantic predictions.

\subsection{Debiased Instance Localization}\label{sec:debiased_inst}

\begin{figure*}[htbp]
    \begin{center}
    \scalebox{1.0}{
        \includegraphics[width=1\linewidth]{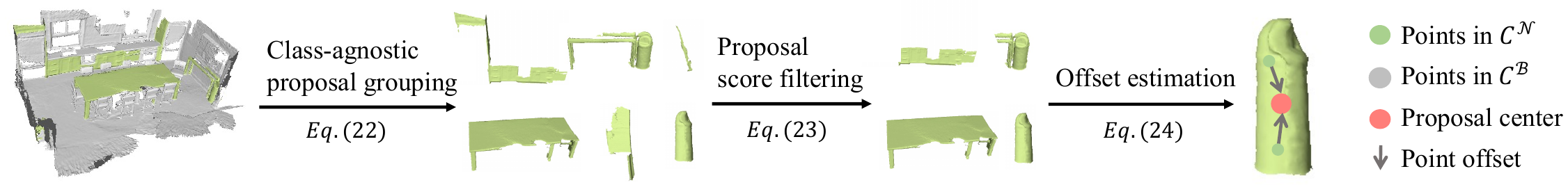}
    }
    \end{center}
    \caption{Debiased Instance Localization. Points belonging to novel categories are grouped together to form candidate proposals. Subsequently, confidence filtering is applied, utilizing proposal scores to exclude potential mis-grouped points. Finally, we estimate proposal centers and point offsets, which serve as pseudo offset supervision signals for novel categories.}
    \label{fig:inst_group}
\end{figure*}

As we discussed in Sec.~\ref{sec:analysis}, the offset branch $\textbf{F}_\text{off}$, trained on base categories, tends to overfit to the instance patterns of base categories and produces poor offset predictions on novel categories. 
This overfitting issue poses a challenge in generating high-quality proposals for novel objects due to unreliable offset predictions.
To address this issue, we propose debiased instance localization (DIL). DIL rectifies the learning bias of $\textbf{F}_\text{off}$ through providing high-quality pseudo-offset supervision signals for unlabeled data containing potential novel objects. It achieves this by candidate proposal grouping, proposal confidence filtering and offset estimation, which is detailed as below.

First, during training, we can group offset-shifted points of base categories effectively by using semantic scores as Eq.~\eqref{eq:inst_basic}. However, unlabeled data do not have prior semantic knowledge. Therefore, we simply treat all points from novel categories belong to one class, which enables to group these offset-shifted points and obtain candidate proposals as follows:
\begin{gather}
    \mathbf{r}^{N}=\mathbf{F}_\text{group}(\mathbf{p}^{N}, \mathbf{o}^{N}),
\end{gather}
where the subscript $N$ indicates unlabeled unseen categories. Fig.~\ref{fig:inst_group} shows examples of the grouped proposals. 
To deal with possible mis-groupings, we further apply confidence filtering based on the proposal score $\textbf{z}$, which estimates the likelihood of each point belonging to a given proposal, as shown in Eq.~\ref{eq:inst_group}. This step helps us filter out points that may have been wrongly grouped and keep only those that belong to the instances with high confidence, as illustrated in Fig.~\ref{fig:inst_group}.
\begin{gather}\label{eq:inst_group}
    \widehat{\mathbf{r}}^{N}=\{p \mid p \in \mathbf{r}^{N} ~\text{and}~ \mathbf{z}^N(p) > \eta \},
\end{gather}
where $\widehat{\textbf{r}}^{N}$ is the refined proposal, $p$ is a point in the proposal, $\textbf{z}^N(p)$ is the score for point $p$ in proposal $\textbf{r}^{N}$, and $\eta$ is the score threshold. After obtaining $\widehat{\textbf{r}}^{N}$, we estimate their centers and then point offsets toward centers as shown in Eq.~\eqref{eq:off} and Fig.~\ref{fig:inst_group}. Those predicted point offsets can serve as pseudo-supervision signals and help the offset branch learn more generalizable features by incorporating more diversity and comprehensiveness. 
\begin{gather}\label{eq:off}
    \mathbf{\widehat{y}}_\text{off}^N=\{p-\text{center}(\widehat{\mathbf{r}}^{N}) \mid p \in \widehat{\mathbf{r}}^{N}\},,
\end{gather}
where $\mathbf{\widehat{y}}_\text{off}^{N}$ denotes the pseudo offset supervision for unlabeled objects, and the center denotes the center estimation of the proposal. 
Therefore, the offset loss in Eq.~\eqref{eq:basic_loss} involves two parts:
\begin{gather}
    \mathcal{L}_\text{off}=\mathcal{L}_\text{off}^B+\mathcal{L}_\text{off}^N, ~~ \mathcal{L}_\text{off}^N = \text{Loss}(\mathbf{o}^N, \widehat{\mathbf{y}}_\text{off}^N)
\end{gather}
where $\mathcal{L}_\text{off}^B$ and $\mathcal{L}_\text{off}^N$ denote offset  prediction loss on base and novel categories, respectively. In this way, the offset branch can be better generalized to unseen categories to benefit open-world instance-level understanding tasks.

Finally, as shown in Fig.~\ref{fig:framework}, the overall training objective of Lowis3D can be written as:

\begin{equation}
    \mathcal{L} = \mathcal{L}_{\text{sem}} + \mathcal{L}_{\text{off}} + \mathcal{L}_{\text{score}} + \mathcal{L}_{\text{cap}}^{\text{all}}+\mathcal{L}_{\text{bi}}.
\end{equation}

\vspace{0.05in}
\subsection{Comparison to Concurrent Work}

Recently, the 3D scene understanding community has made concurrent efforts to leverage visual-language (VL) foundation models. OpenScene~\cite{Peng2023OpenScene} uses 2D open-vocabulary segmentors such as LSeg~\cite{li2022languagedriven} and OpenSeg~\cite{ghiasi2022scaling} to extract pixel-level embeddings aligned with 3D points, enabling 3D semantic-level understanding through techniques such as zero-shot fusion or feature distillation. Similarly, CLIP2Scene~\cite{chen2023clip2scene} employs MaskCLIP~\cite{zhou2022maskclip} to obtain pixel-aligned features for annotation-free and label-efficient scene understanding. ConceptFusion~\cite{jatavallabhula2023conceptfusion} and CLIP-FO3D~\cite{zhang2023clip} further explore acquiring pixel-aligned knowledge through dense region-level feature extraction using CLIP~\cite{radford2021learning} and multi-view feature fusion. These methods rely on semantic-aware visual features to guide 3D scene understanding. In contrast, Lowis3D adopts a different approach by utilizing pure language supervision to inject rich semantics into the 3D network, building an efficient training and inference pipeline for open-world scene understanding. Moreover, these existing methods may face difficulties in performing instance localization due to the lack of objectness information, which is specifically addressed by Lowis3D. This unique instance localization aspect of our approach broadens its potential applications in fields such as robotics or autonomous driving, where the detection and tracking of unseen objects is desired.

Besides, there have been attempts to perform instance segmentation such as CLIP2~\cite{zeng2023clip2} and RegionPLC~\cite{yang2023regionplc}. They use region-level supervision signals that encode objectness information from image patches or object proposals to perform instance segmentation. While their main goal is to inject fine-grained semantics into the 3D network to facilitate object localization, our Lowis3D focuses on a different aspect by correcting the network bias to learn a more general localization branch. Importantly, we empirically show that these two work streams can work together effectively to improve instance segmentation, as shown in Table~\ref{tab:comb_region} for more details.

\begin{table*}[htbp]
    \caption{Open-world 3D instance segmentation results on ScanNet and S3DIS in terms of hAP$_{50}$, mAP$_{50}^\mathcal{B}$ and mAP$_{50}^\mathcal{N}$. $\mathcal{C}^N$ prior refers to whether novel category names $\mathcal{C}^N$ are known during training. Best open-world results are highlighted in \textbf{bold}.}

    \label{tab:inst_seg}
    
    \centering
    \setlength\tabcolsep{2pt}
    \scalebox{1.0}{
        \begin{tabular}{c|c|ccc|ccc|ccc|ccc|ccc}
            \bottomrule[1pt]
            \multirow{3}{*}{Method} & \multirow{3}{*}{$\mathcal{C}^N$ prior} & \multicolumn{9}{c|}{ScanNet} & \multicolumn{6}{c}{S3DIS} \\
            \cline{3-17}
            & & \multicolumn{3}{c|}{B13/N4} & \multicolumn{3}{c|}{B10/N7} & \multicolumn{3}{c|}{B8/N9} & \multicolumn{3}{c|}{B8/N4} & \multicolumn{3}{c}{B6/N6} \\
            \cline{3-17}
            & & hAP$_{50}$ & mAP$_{50}^\mathcal{B}$ & mAP$_{50}^\mathcal{N}$ & hAP$_{50}$ & mAP$_{50}^\mathcal{B}$ & mAP$_{50}^\mathcal{N}$ & hAP$_{50}$ & mAP$_{50}^\mathcal{B}$ & mAP$_{50}^\mathcal{N}$ & hAP$_{50}$ & mAP$_{50}^\mathcal{B}$ & mAP$_{50}^\mathcal{N}$ & hAP$_{50}$ & mAP$_{50}^\mathcal{B}$ & mAP$_{50}^\mathcal{N}$ \\
            \hline
            OV-SoftGroup~\cite{li2022languagedriven} & $\times$ & 05.1 & 57.9 & 02.6 & 02.0 & 50.7 & 01.0 & 02.4 & 59.4 & 01.2 & 00.5 & 58.3 & 00.3 & 01.1 & 41.4 & 00.5 \\ 
            PLA~\cite{ding2022language} & $\times$ & 55.5 & 58.5 & 52.9 & 31.2 & 54.6 & 21.9 & 35.9 & 63.1 & 25.1 & 15.0 & 59.0 & 08.6 & \textbf{16.0} & 46.9 & 09.8 \\
            \hline
            Lowis3D& $\times$ & \textbf{59.1} & 58.6 & \textbf{59.6} & \textbf{40.0} & 55.5 & \textbf{31.2} & \textbf{47.6} & \textbf{63.5} & \textbf{38.1} & \textbf{22.3} & 58.7 & \textbf{13.8} & 24.2 & \textbf{51.8} & \textbf{15.8} \\
            \hline
            Fully-Sup. & \checkmark & 64.5 & 59.4 & 70.5 & 62.5 & 57.6 & 62.0 & 62.0 & 65.1 & 62.0 & 57.6 & 60.8 & 54.6 & 57.4 & 50.0 & 67.5 \\
            \toprule[0.8pt]
        \end{tabular}
     }
\end{table*}

\begin{table*}[htbp]
    \caption{Open-world 3D panoptic segmentation results on nuScenes in terms of panoptic quality (hPQ, PQ$^\mathcal{B}$, PQ$^\mathcal{N}$), recognition quality (hRQ, RQ$^\mathcal{B}$, RQ$^\mathcal{N}$) and segmentation quality (hSQ, SQ$^\mathcal{B}$, SQ$^\mathcal{N}$).}
    \label{tab:panoptic_seg}
    \centering
    \setlength\tabcolsep{3pt}
    \scalebox{1.0}{
        \begin{tabular}{c|c|ccc|ccc|ccc|ccc|ccc|ccc}
            \bottomrule[1pt]
            \multirow{3}{*}{Method} & \multirow{3}{*}{$\mathcal{C}^N$ prior} & \multicolumn{18}{c}{nuScenes} \\
            \cline{3-20}
            & & \multicolumn{9}{c|}{B12/N3} & \multicolumn{9}{c}{B10/N5} \\
            \cline{3-20}
            & & hPQ & PQ$^\mathcal{B}$ & PQ$^\mathcal{N}$ & hRQ & RQ$^\mathcal{B}$ & RQ$^\mathcal{N}$ & hSQ & SQ$^\mathcal{B}$ & SQ$^\mathcal{N}$ & hPQ & PQ$^\mathcal{B}$ & PQ$^\mathcal{N}$ & hRQ & RQ$^\mathcal{B}$ & RQ$^\mathcal{N}$ & hSQ & SQ$^\mathcal{B}$ & SQ$^\mathcal{N}$ \\
            \hline
            OV-SoftGroup~\cite{li2022languagedriven} & $\times$ & 00.1 & 46.4 & 00.0 & 00.2 & 53.9 & 00.1 & 00.0 & 43.3 & 00.0 & 00.0 & 40.9 & 00.0 & 00.0 & 47.3 & 00.0 & 31.6 & 74.7 & 20.0 \\             
            PLA~\cite{ding2022language} & $\times$ & 30.8 & 48.4 & 22.6 & 34.9 & 56.5 & 25.3 & 77.3 & 77.2 & 77.5 & 12.3 & 45.1 & 07.1 & 14.7 & 51.6 & 08.6 & 64.8 & \textbf{76.0} & 56.5 \\
            \hline
            Lowis3D & $\times$ & \textbf{43.4} & \textbf{49.6} & \textbf{38.6} & \textbf{49.4} & \textbf{58.1} & \textbf{42.9} & \textbf{80.1} & \textbf{77.3} & \textbf{83.1} & \textbf{14.7} & \textbf{45.4} & \textbf{08.8} & \textbf{17.1} & \textbf{52.7} & \textbf{10.2} & \textbf{75.3} & 75.8 & \textbf{74.9} \\
            \hline
            Fully-Sup. & \checkmark &  54.7 & 48.0 & 63.5 & 61.8 & 55.9 & 69.0 & 84.3 & 76.5 & 92.2 & 52.6 & 45.0 & 63.4 & 60.3 & 51.8 & 72.0 & 81.4 & 75.3 & 88.5 \\
            \toprule[0.8pt]
        \end{tabular}
     }
\end{table*}

\begin{table*}[htbp]
    \caption{Oopen-world 3D semantic segmentation results on ScanNet and S3DIS in terms of hIoU, mIoU$^\mathcal{B}$ and mIoU$^\mathcal{N}$.  PLA (w/o Cap.) refers to the model trained without using point-language pairs as supervision. Notice that Lowis3D uses the same semantic module as PLA, so their semantic performance are identical. }
    \label{tab:sem_seg}
    \centering
    \setlength\tabcolsep{2pt}
    \scalebox{1.0}{
        \begin{tabular}{c|c|ccc|ccc|ccc|ccc|ccc}
            \bottomrule[1pt]
            \multirow{3}{*}{Method} & \multirow{3}{*}{$\mathcal{C}^N$ prior}  & \multicolumn{9}{c|}{ScanNet} & \multicolumn{6}{c}{S3DIS} \\
            \cline{3-17}
            & & \multicolumn{3}{c|}{B15/N4} & \multicolumn{3}{c|}{B12/N7} & \multicolumn{3}{c|}{B10/N9} & \multicolumn{3}{c|}{B8/N4} & \multicolumn{3}{c}{B6/N6} \\
            \cline{3-17}
            & & hIoU & mIoU$^\mathcal{B}$ & mIoU$^\mathcal{N}$ & hIoU & mIoU$^\mathcal{B}$ & mIoU$^\mathcal{N}$ & hIoU & mIoU$^\mathcal{B}$ & mIoU$^\mathcal{N}$ & hIoU & mIoU$^\mathcal{B}$ & mIoU$^\mathcal{N}$ & hIoU & mIoU$^\mathcal{B}$ & mIoU$^\mathcal{N}$ \\
            \hline
            OV-SparseConvNet~\cite{li2022languagedriven} & $\times$ & 00.0 & 64.4 & 00.0 & 00.9 & 55.7 & 00.1 & 01.8 & 68.4 & 00.9 & 00.1 & 49.0 & 00.1 & 00.0 & 30.1 & 00.0 \\ 
            3DTZSL~\cite{cheraghian2020transductive} & \checkmark & 10.5 & 36.7 & 06.1 & 03.8 & 36.6 & 02.0 & 07.8 & 55.5 & 04.2 & 08.4 & 43.1 & 04.7 & 03.5 & 28.2 & 01.9 \\
            3DGenZ~\cite{michele2021generative} & \checkmark & 20.6 & 56.0 & 12.6 & 19.8 & 35.5 & 13.3 & 12.0 & 63.6 & 06.6 & 08.8 & 50.3 & 04.8 & 09.4 & 20.3 & 06.1 \\
            \hline
            PLA (w/o Cap.) & $\times$  & 39.7 & \textbf{68.3} & 28.0 & 24.5 & \textbf{70.0} & 14.8 & 25.7 & 75.6 & 15.5 & 13.0 & 58.0 & 07.4 & 12.2 & 54.5 & 06.8\\
            PLA / Lowis3D & $\times$ & \textbf{65.3} & \textbf{68.3} & \textbf{62.4} & \textbf{55.3} & 69.5 & \textbf{45.9} & \textbf{53.1} & \textbf{76.2} & \textbf{40.8} & \textbf{34.6} & \textbf{59.0} & \textbf{24.5} & \textbf{38.5} & \textbf{55.5} & \textbf{29.4} \\
            \hline
            Fully-Sup. & \checkmark & 73.3 & 68.4 & 79.1 & 70.6 & 70.0 & 71.8 & 69.9 & 75.8 & 64.9 & 67.5 & 61.4 & 75.0 & 65.4 & 59.9 & 72.0 \\
            \toprule[0.8pt]
        \end{tabular}
     }
  
\end{table*}

\section{Experiments}
\subsection{Basic Setups}
\noindent\textbf{Datasets.}
To thoroughly validate the effectiveness of Lowis3D, we conduct experiments on two indoor datasets, \ie ScanNet~\cite{dai2017scannet} annotated in 20 classes,  S3DIS~\cite{armeni20163d} with 13 classes, for both semantic and instance segmentation tasks. Additionally, we evaluate Lowis3D on an outdoor dataset, \ie nuScenes~\cite{fong2022panoptic} consisting of 16 classes on panoptic segmentation.

\vspace{0.05in}
\noindent\textbf{Category Partitions.}
As there are no standard open-world partitions available for the ScanNet, S3DIS, and nuScenes datasets, we create our own open-world benchmark with multiple base/novel partitions. 
To avoid confusion in the models, we disregard the ``otherfurniture'' class in ScanNet, the ``clutter'' class in S3DIS and the ``other\_flat'' class in nuScenes since they lack precise semantic meanings and can encompass any semantic classes. Besides, for instance segmentation, we exclude two background classes and randomly divide the rest 17 classes into 3 base/novel partitions in ScanNet: B13/N4, B10/N7 and B8/N9. Here, B13/N4 indicates 13 base categories and 4 novel categories. For semantic segmentation, we add the two background classes to base categories and thus obtain B15/N4, B12/N7 and B10/N9 partitions. 
Regarding the S3DIS dataset, we randomly shuffle the remaining 12 classes into 2 base/novel splits: B8/N4, B6/N6, for semantic and instance segmentation. For nuScenes~\cite{fong2022panoptic} panoptic segmentation, we split the rest 15 categories into B12/N3 and B10/N5 partitions.
Specific category splits can be found in the Appendix~\ref{sec:dataset}.

\vspace{0.05in}
\noindent\textbf{Metrics.} 
We utilize the widely adopted metrics of mean intersection over union (mIoU), mean average precision under 50\% IoU threshold (mAP$_{50}$) as evaluation metrics for semantic segmentation and instance segmentation, respectively. Besides, we apply panoptic quality (PQ), which can be decomposed to segmentation quality (SQ) and recognition quality (RQ) as metrics for panoptic segmentation. 
These evaluation metrics are computed on base and novel categories, with the superscripts of $\mathcal{B}$ and $\mathcal{N}$ (\eg mIoU$^\mathcal{B}$), respectively. 
Furthermore, we use the harmonic metric such as harmonic IoU (hIoU) as major indicators for open-world tasks following popular zero-shot learning works~\cite{xian2019semantic,xu2021simple} to consider category partition between base and novel classes.

\vspace{0.05in}
\noindent\textbf{Network Architectures.} We employ the popular and high-performance sparse convolutional UNet~\cite{graham20183d,choy20194d} as 3D encoder $\text{F}_{\text{3D}}$, the text encoder of CLIP as $\text{F}_{\text{text}}$, fully-connected layers with batch normalization~\cite{ioffe2015batch} and ReLU~\cite{nair2010rectified} as VL adapter $\text{F}_{\theta}$, an UNet decoder as binary head $\text{F}_{\text{b}}$. Additionally, we adopt the state-of-the-art instance segmentation network SoftGroup~\cite{vu2022softgroup} for proposal grouping $\text{F}_\text{off}$ and scoring $\text{F}_\text{score}$. We set voxel size as 0.02 for indoor datasets and 0.1 for outdoor datasets.

\vspace{0.05in}
\noindent\textbf{Baseline Methods.}
For instance and panoptic segmentation, we employ \textbf{OV-Softgroup} as a baseline. Given that instance-level open-world 3D scene understanding is still in its infancy, there are currently no other proper methods for direct comparison.
For semantic segmentation, in addition to the \textbf{OV-SparseConvNet} mentioned in Sec.\ref{sec:text_embed_cls}, we also re-produce two 3D zero-shot learning approach, namely  \textbf{3DGenZ}~\cite{michele2021generative} and \textbf{3DTZSL}~\cite{cheraghian2020transductive} with task-tailored modifications.
Specifically, for 3DGenZ~\cite{michele2021generative}, instead of training the model on samples containing only base classes, we train it on the entire training dataset, where points belonging to novel classes are ignored during optimization. We omit the calibrated stacking component of 3DGenZ, as it has shown only minor performance gains in our implementations. Regarding 3DTZSL~\cite{cheraghian2020transductive}, originally designed for object classification, we extend it for semantic segmentation by adapting it to learn with triplet loss at the point level instead of the sample level. The projection net of 3DTZSL is implemented using one or two fully-connected layers with the Tanh activation function, as described in the original paper.
Furthermore, these methods are reproduced using the same 3D backbone and CLIP text embeddings to ensure fair comparisons.

\vspace{0.05in}
\noindent\textbf{Implementation Details.} 
In the indoor experiments, we train for 19,216 iterations on ScanNet and 4,080 iterations on S3DIS for the semantic segmentation task. For instance segmentation, we train for 22,520 iterations on ScanNet and 9,160 iterations on S3DIS. The initial learning rate is set to 0.004 with cosine decay for the learning rate schedule.
For the outdoor panoptic experiments on nuScenes, we train for 61,600 iterations. The learning rate is initialized as 0.006 with polynomial decay. We employ the AdamW~\cite{loshchilov2017decoupled} optimizer and run all experiments with a batch size of 32 on either 8 NVIDIA A100 or NVIDIA V100 GPUs. Regarding entity-level captions, we apply a filtering process on $\langle \mathbf{\hat{p}^e},\mathbf{t}^e \rangle$ pairs to ensure that the point set $\mathbf{\hat{p}^e}$ contains only a few entities and remains small enough. Specifically, we set the minimal points $\gamma$ as 100 and control the maximum point ratio $\delta$ to 0.3.
As for the caption loss, in the indoor experiments on ScanNet, we set the weights $\alpha_1$, $\alpha_2$ and $\alpha_3$ as 0, 0.05 and 0.05, respectively, for the  scene-level loss $\mathcal{L}^{s}_\text{cap}$, view-level loss $\mathcal{L}^{v}_\text{cap}$ and entity-level loss $\mathcal{L}^{e}_\text{cap}$. For S3DIS, we set the weights $\alpha_1$, $\alpha_2$, and $\alpha_3$ as 0, 0.08, and 0.02 separately. 
In the outdoor experiments, since each outdoor scene contains only 6 images, the scene-level coverage may be limited, and acquiring entity-level captions is challenging due to the high similarity between images. Thus, we set $\alpha_1$, $\alpha_2$, and $\alpha_3$ as 0, 0.1, and 0, respectively.

\subsection{Main Results}\label{sec:sem_seg}

\noindent\textbf{3D Instance Segmentation.}
Table~\ref{tab:inst_seg} clearly demonstrates the remarkable superiority of our method over the OV-SoftGroup baseline. We achieve an improvement of 38.0\% $\sim$ 54.0\% in hAP$_{50}$ on ScanNet and 21.8\% $\sim$ 23.1\% on S3DIS, across different base/novel partitions. This significant performance boost highlights the effectiveness of our contrastive point-language training in enabling the 3D backbone to learn both semantic attributes and instance localization information from rich captions.  Additionally, Compared with PLA, our Lowis3D further achieves an additional performance gain of 3.6\% $\sim$ 11.7\% hAP$_{50}$ across different partitions on two datasets.  This further confirms the substantial enhancement in localization generalization on novel categories brought about by our debiased instance localization module. It is worth noting that the improvement for the S3DIS dataset is smaller compared to ScanNet. This can be attributed to the smaller number of training samples in S3DIS (only 271 scenes) and the fewer point-caption pairs available due to the limited overlapping regions between images and 3D scenes in this dataset.

\vspace{0.05in}
\noindent\textbf{3D Panoptic Segmentation.} While Lowis3D has demonstrated remarkable performance in open-world scene understanding for indoor scenes, we also conduct validation experiments on outdoor LiDAR-scanned scenes, specifically focusing on the panoptic segmentation task.
As shown in Table~\ref{tab:panoptic_seg}, Lowis3D achieves a remarkable improvement in hPQ, with a gain of 14.7\% $\sim$ 43.3\% over the OV-SoftGroup baseline. Moreover, both hRQ and hSQ show notable improvements of 17.1\% $\sim$ 49.7\% and 43.7\% $\sim$ 80.1\%, respectively. These results demonstrate the coherent recognition and localization capabilities of Lowis3D. Besides, Lowis3D surpasses PLA by a considerable margin of 2.4\% $\sim$ 12.6\%, further validating that the debiased instance localization greatly enhances its general objectness comprehending ability in the open world. 
Overall, these findings demonstrate the effectiveness of Lowis3D in achieving impressive performance in outdoor panoptic segmentation tasks, reinforcing its strengths in open-world scene understanding across various scenarios.

\vspace{0.05in}
\noindent\textbf{3D Semantic Segmentation.} To more straightly show the open-world semantic recognition ability of our method, we compare Lowis3D with other baselines. The results presented in Table~\ref{tab:sem_seg} clearly demonstrate the superiority of our method compared to the OV-SparseConvNet~\cite{li2022languagedriven} baseline, with significant improvements around 51.3\% $\sim$ 65.3\% and 34.5\% $\sim$ 38.5\% hIoU across different partitions on ScanNet and S3DIS, respectively, showcasing the model's outstanding open-world capability.
Our method also outperforms prior zero-shot methods  3DGenZ~\cite{michele2021generative} and 3DTZSL~\cite{cheraghian2020transductive}, despite the advantage these methods have of knowing the novel category names during training. Our method achieves 35.5\% $\sim$ 54.8\% improvements in terms of hIoU among various partitions on ScanNet. 
In particular, PLA / Lowis3D largely surpasses its counterpart without language supervision (\ie PLA (w/o Cap.)) by 25.6\% $\sim$ 30.8\% hIoU and 21.6\% $\sim$ 26.3\% hIoU on ScanNet and S3DIS, respectively. The consistent performance of our method across different base/novel partitions and datasets emphasizes its effectiveness and robustness, regardless of the specific configuration of the data. This makes it a highly adaptable and reliable model for a wide range of 3D scene understanding tasks.

\vspace{0.05in}
\noindent\textbf{Self-Bootstrap with Novel Category Prior.} 
In addition to our main method, we also present a simple variant that leverages novel category priors in a self-training fashion, similar to existing zero-shot methods such as 3DGenZ~\cite{michele2021generative} and 3DTZSL~\cite{cheraghian2020transductive}. This variant allows our model to access novel category names during training without any human annotation. As shown in Table~\ref{tab:self_train}, Lowis3D (w/ self-train) obtains 3.1\% $\sim$ 6.6\% gains for instance segmentation on ScanNet across various partitions. This demonstrates that our model can further self-bootstrap its open-world capability and extend its vocabulary size without relying on any manual annotation. 

\begin{table}[htbp]
    \caption{Self-training results of instance segmentation on ScanNet wth novel category names as prior in terms of hAP$_{50}$ / mAP$_{50}^\mathcal{B}$ / mAP$_{50}^\mathcal{N}$.}
    \label{tab:self_train}
    
    \centering
    \setlength\tabcolsep{1pt}
    \scalebox{0.94}{
        \begin{tabular}{c|c|c|c|c}
            \bottomrule[1pt]
            \multirow{2}{*}{Method} & \multirow{2}{*}{\makecell{$\mathcal{C}^N$ \\ prior}} & \multicolumn{3}{c}{hAP$_{50}$ / mAP$_{50}^\mathcal{B}$ / mAP$_{50}^\mathcal{N}$} \\
            \cline{3-5}
            & & \multicolumn{1}{c|}{B13/N4} & \multicolumn{1}{c|}{B10/N7} & \multicolumn{1}{c}{B8/N9}  \\
            \hline
            Lowis3D& $\times$ & 59.1 / 58.6 / 59.6 & 40.0 / 55.5 / 31.2 & 47.6 / 63.5 / 38.1 \\
            Lowis3D (w/ self-train) & \checkmark & \textbf{62.2} / \textbf{58.9} / \textbf{65.8} & \textbf{46.6} / \textbf{56.7} / \textbf{39.6} & \textbf{51.6} / \textbf{64.9} / \textbf{42.7} \\
            \toprule[0.8pt]
        \end{tabular}
     }
\end{table}

\section{Ablation Studies}\label{sec:ablation}
In this section, we examine the key components of our open-world instance-level scene understanding framework through in-depth ablation studies, which covers two major aspects -- semantic recognition and instance localization. 
Experiments are conducted on ScanNet B13/N4 partition by default (\ie B13/N4 for instance segmentation and B15/N4 for semantic segmentation). The default setting is marked in {\colorbox{mygray}{gray}} and the best results are highlighted in bold.

\vspace{0.05in}
\noindent\textbf{Component Analysis.}
We investigate the effectiveness of our proposed modules, \ie the binary calibration module, three coarse-to-fine point-language supervision manners and the debiased instance localization.
As shown in Table~\ref{tab:component}, the adoption of the binary calibration module for semantic calibration demonstrates significant improvements over the OV-SparseConvNet baseline, achieving a 39.8\% increase in hIoU for semantic segmentation. Similarly, compared to the OV-SoftGroup baseline, the binary calibration module leads to a substantial 15.9\% improvement in hAP$_{50}$ for instance segmentation.
Such substantial performance boosts on both base and novel classes validates the effectiveness of the binary calibration module in rectifying semantic scores and improving the overall segmentation accuracy.

As for the point-language association manners, they all considerably  improve the results by a significant margin of 14.8\% $\sim$ 23.8\% hIoU and 31.8\% $\sim$ 35.6\% hAP$_{50}$ on semantic and instance segmentation, respectively. Among the three association manners, entity-level language supervision demonstrates the best performance, highlighting the importance of fine-grained caption-point correspondence in constructing effective point-caption pairs. This finding suggests that capturing detailed and specific information at the object instance level is crucial for improving segmentation accuracy. It should be noted that when we combine three types of captions with the same loss weight, the model does not always yield boosts in all scenarios, potentially attributed to the challenges of simultaneously optimizing various caption losses of different granularities.

Regarding debiased instance localization, it greatly lifts the instance segmentation results by 3.6\% hAP$_{50}$ and 6.7\% AP$_{50}^\mathcal{N}$. It demonstrates that it significantly enhances the robustness and generalization of proposal grouping, thereby improving the instance localization capabilities with respect to novel categories. This finding confirms that the objectness bias towards base patterns can be accurately rectified by learning from more unseen and diverse samples.

The combination of the proposed modules ultimately leads to an overall better performance in 3D scene understanding tasks, including semantic recognition and instance localization.

\begin{table}[htbp]
    \caption{Component analysis in terms of hIoU / mIoU$^\mathcal{B}$ /mIoU$^\mathcal{N}$ and hAP$_{50}$ / mAP$_{50}^\mathcal{B}$ / mAP$_{50}^\mathcal{N}$. Binary denotes binary head calibration. Cap$^s$, Cap$^v$ and Cap$^e$ denotes scene-level, view-level and entity-level caption supervision, respectively. DIL denotes debiased instance localization.}
    \label{tab:component}
    \centering
    \setlength\tabcolsep{0.5pt}
    \scalebox{0.95}{
        \begin{tabular}{c|c|c|c|c|c|c}
            \bottomrule[1pt]
            \multicolumn{5}{c|}{Components} & \multirow{2}{*}{hIoU / mIoU$^\mathcal{B}$ /mIoU$^\mathcal{N}$} & \multirow{2}{*}{hAP$_{50}$ / mAP$_{50}^\mathcal{B}$ / mAP$_{50}^\mathcal{N}$} \\
            \cline{1-5}
            Binary & Cap$^s$ & Cap$^v$ & Cap$^e$ & DIL & & \\
            \hline
            & & & & & 00.0 / 64.4 / 00.0 & 05.1 / 57.9 / 02.6\\
            \hline
            \checkmark & & & & & 39.8 / \textbf{68.5} / 28.1 & 21.0 / \textbf{59.6} / 12.8 \\
            \hline
            \checkmark & \checkmark & & & & 54.6 / 67.9 / 45.7 & 52.8 / 57.8 / 36.6 \\
            \checkmark & & \checkmark & & & 61.3 / \textbf{68.5} / 55.5 & 55.9 / 58.9 / 53.3 \\
            \checkmark & & & \checkmark & & 63.6 / 67.8 / 60.0 &  \textbf{56.6} / 59.0 / \textbf{54.4} \\
            \hline
            \checkmark & \checkmark & \checkmark & & & 61.9 / 68.1 / 56.8 & 54.9 / 59.5 / 51.0 \\
            \checkmark &  & \checkmark & \checkmark & & \textbf{65.3} / 68.3 / \textbf{62.4} & 55.5 / 58.5 / 52.9 \\
            \checkmark & \checkmark & \checkmark & \checkmark & & 64.6 / 69.0 / 60.8 & 54.5 / 58.2 / 51.4 \\
            \hline
            \checkmark & \checkmark & \checkmark &  &  \checkmark& \cellcolor{mygray}\textbf{65.3} / 68.3 / \textbf{62.4} & \cellcolor{mygray}\textbf{59.1} / 58.6 / \textbf{59.6} \\
            \toprule[0.8pt]
        \end{tabular}
    }
\end{table}

\begin{figure*}[htbp]
    \begin{center}
    \scalebox{1.0}{
        \includegraphics[width=1\linewidth]{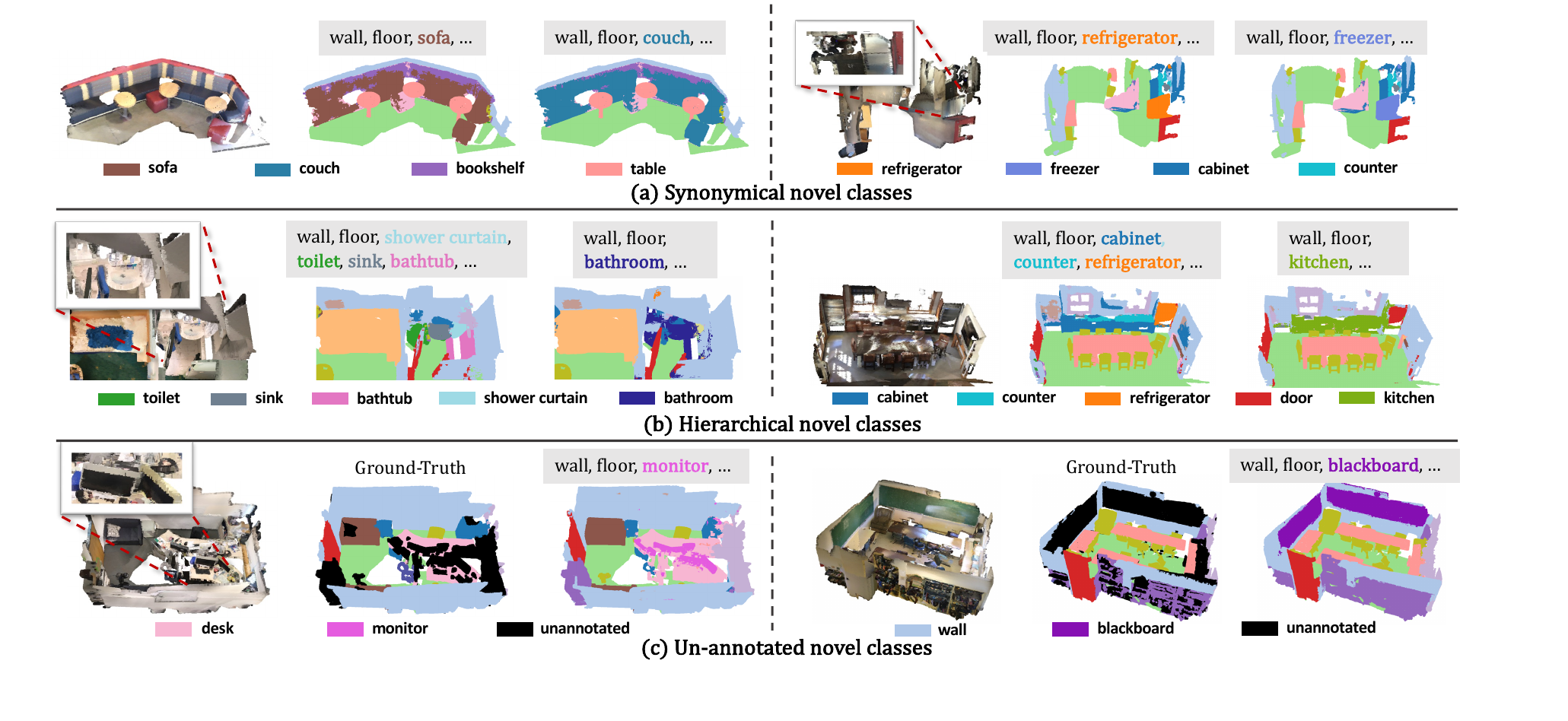}
    }
    \end{center}

    \caption{Qualitative examples of identifying out-of-vocabulary categories. (a) shows the results of identifying synonymical categories. (b) presents the segmentation results on abstract concepts. (c) illustrates the results of segmenting unannotated classes.}
    \label{fig:vis}
\end{figure*}

\vspace{0.05in}
\noindent\textbf{Caption Composition Analysis.} 
We delve into a comprehensive exploration of the types of words that predominantly contribute to the open-world capability, given that captions can composite various elements such as entities (\eg sofa), their relationships (\eg spatial relation), and attributes (\eg color and texture). Table~\ref{tab:caption_components} illustrates that when we retain only entity phrases within the caption, variant (a) even surpasses the full-caption variant. Furthermore, when we only keep the entities in the captions that precisely align with category names, we observe a considerable over $13\%$ mIoU decline in the resultant variant (b) in terms of novel categories. This suggests the importance of a diverse vocabulary that expands the semantic scope in maintaining the efficacy of captions. Moreover, even though variant (c) integrates both accurate base and novel label names within the captions, its performance marginally lags behind our foundation-model-generated captions. This demonstrates that existing foundation models are powerful enough to provide promising supervisions.


\begin{table}[htbp]
    \caption{Ablation of caption compositions in terms of hIoU / mIoU$^\mathcal{B}$ / mIoU$^\mathcal{N}$.}
    \label{tab:caption_components}
    \centering
    \setlength\tabcolsep{5pt}
    \scalebox{1.0}{
        \begin{tabular}{l|c}
            \bottomrule[1pt]
            Caption Composition & hIoU / mIoU$^\mathcal{B}$ / mIoU$^\mathcal{N}$ \\
            \hline
            (a) keep only entities & \textbf{65.7} / \textbf{69.0} / \textbf{62.7} \\
            (b) keep only label names & 57.6 / 68.5 / 49.6 \\
            (c) ground-truth label names & 64.8 / 68.1 / 61.9 \\
            (d) full caption & \cellcolor{mygray}65.3 / 68.3 / 62.4\\ 
            \toprule[0.8pt]
        \end{tabular}
    }
\end{table}

\vspace{0.05in}
\noindent\textbf{Text Encoder Selection.}
Here, we examine different text encoders $\text{F}_\text{text}$ for extracting caption and category embeddings. As illustrated in Table~\ref{tab:text_encoder}, the text encoder of CLIP~\cite{radford2021learning}, pre-trained on vision-language tasks,  exhibits a performance superior by over 7\% in mIoU$^\mathcal{N}$ compared to BERT~\cite{devlin2018bert} and GPT2~\cite{radford2019language}, both of which are exclusively pre-trained on language modality. This  evidences that a text encoder which is aware of visual elements can provide superior semantic embedding for 3D-language tasks. This is potentially because 3D tasks also utilize information such as texture, shape, and RGB values for recognition, similar to image-based tasks.

\begin{table}[htbp]
    \caption{Ablation of text encoders for extracting text embeddings in terms of hIoU / mIoU$^\mathcal{B}$ / mIoU$^\mathcal{N}$.}
    \label{tab:text_encoder}
    \centering
    \setlength\tabcolsep{2pt}
    \scalebox{0.92}{
        \begin{tabular}{c|c|c|c}
            \bottomrule[1pt]
            Text Encoder & BERT~\cite{devlin2018bert} & GPT2~\cite{radford2019language} & CLIP~\cite{radford2021learning} \\
            \hline
            hIoU / mIoU$^\mathcal{B}$ / mIoU$^\mathcal{N}$ & 61.2 / 68.7 / 55.2 & 61.0 / \textbf{69.1} / 54.6 & \cellcolor{mygray}\textbf{65.3} / 68.3 / \textbf{62.4}\\
            \toprule[0.8pt]
        \end{tabular}
    }
\end{table}

\vspace{0.05in}
\noindent\textbf{Foundation Model for Image Captioning.}
Indeed, the choice of the foundation model for image captioning can have a significant impact on open-world performance. In our main experiments, we use GPT-ViT2, which is a popular open-source image-captioning model available on the HuggingFace platform. 
Nevertheless, as demonstrated in Table~\ref{tab:foundation_model}, the recent cutting-edge foundation model OFA~\cite{wang2022ofa} consistently outperforms GPT-ViT2 across all four partitions. 
This indicates that the performance of our method can be further enhanced when paired with more robust and advanced foundation models.

\begin{table}[htbp]
    \caption{Investigation of VL foundation model for image captioning in terms of hIoU / mIoU$^\mathcal{B}$ / mIoU$^\mathcal{N}$.}
    \centering
    \setlength\tabcolsep{3pt}
    \scalebox{1.0}{
        \begin{tabular}{l|c|c|c}
            \bottomrule[1pt]
            \multirow{2}{*}{model} & \multicolumn{3}{c}{hIoU / mIoU$^\mathcal{B}$ / mIoU$^\mathcal{N}$} \\
            \cline{2-4}
            & B15/N4 & B12/N7 & B10/N9 \\
            \hline
            ViT-GPT2~\cite{vit-gpt2} & \cellcolor{mygray}65.3 / \textbf{68.3} / 62.4 & \cellcolor{mygray}55.3 / 69.5 / 45.9 & \cellcolor{mygray}53.1 / \textbf{76.2} / 40.8 \\
            OFA~\cite{wang2022ofa} & \textbf{65.6} / \textbf{68.3} / \textbf{63.1} & \textbf{57.5} / \textbf{69.8} / \textbf{48.9} & \textbf{56.6} / 75.9 / \textbf{45.1} \\
            \toprule[0.8pt]
        \end{tabular}
    }
    \label{tab:foundation_model}
\end{table}

\vspace{0.05in}
\noindent\textbf{Combination of Three Caption Supervisions.} The combination of three types of captions can leads to a 0.6\% increase in hIoU compared to our default setting, as shown in Table~\ref{tab:loss_weight}. However, striking the right balance between these captions demands sophisticated loss trade-off techniques, which may not be generally applicable across different datasets and partitions. Thus, we do not use the scene-level language supervision in the main experiments for the sake of generalization. Future research on effectively combining caption supervisions presents an interesting avenue for future investigation.

\begin{table}[htbp]
    \caption{Ablation for caption loss weights in terms of hIoU / mIoU$^\mathcal{B}$ / mIoU$^\mathcal{N}$.}
    \centering
    \setlength\tabcolsep{4pt}
    \scalebox{1.0}{
        \begin{tabular}{ccc|c}
            \bottomrule[1pt]
            $\alpha_1$(scene) & $\alpha_2$(view) & $\alpha_3$(entity) & hIoU / mIoU$^\mathcal{B}$ / mIoU$^\mathcal{N}$ \\
            \hline
            0.000 & 0.050 & 0.050 & \cellcolor{mygray} 65.3 / 68.3 / 62.4\\
            0.033 & 0.033 & 0.033 & 64.6 / \textbf{69.0} / 60.8 \\
            0.010 & 0.045 & 0.045 & \textbf{65.9} / 68.2 / \textbf{63.8} \\
            \toprule[0.8pt]
        \end{tabular}
    }
    \label{tab:loss_weight}
\end{table}

\vspace{0.05in}
\noindent\textbf{Debiased Instance Localization.}
We assess the effectiveness of our debiased instance localization module, which mitigates learning bias towards base categories. As highlighted in Table~\ref{tab:offset}, the mean absolute error (mAE) on novel classes is significantly reduced by approximately 45.0\%, while the average recall (AR) for proposals improves by 15.4\%. The metrics for base classes remain unaffected. This verifies that our debiased instance localization module significantly enhances the generalizability of offset learning and substantially boosts the ability to localize novel objects.

\begin{table}[htbp]
\caption{Ablation for debiased instance localization on ScanNet B13/N4 in terms of proposal hAR / AR$^\mathcal{B}$ / AR$^\mathcal{N}$ and offset hAE / mAE$^\mathcal{B}$ / mAE$^\mathcal{N}$.}
    \centering
    \setlength\tabcolsep{5pt}
    \scalebox{1.0}{
        \begin{tabular}{c|c|c}
            \bottomrule[1pt]
            DIL  & offset hAE / AE$^\mathcal{B}$ / AE$^\mathcal{N}$ ($\downarrow$) & hAR / AR$^\mathcal{B}$ / AR$^\mathcal{N}$ ($\uparrow$)\\
            \hline
            $\times$ & 0.50 / 0.39 / 0.69 & 44.7 / \textbf{47.4} / 42.3\\
            $\checkmark$ & \cellcolor{mygray}\textbf{0.43} / \cellcolor{mygray}\textbf{0.38} / \cellcolor{mygray}\textbf{0.48} & \cellcolor{mygray}\textbf{47.9} / \cellcolor{mygray}47.1 / \cellcolor{mygray}\textbf{48.8}\\
            \toprule[0.8pt]
        \end{tabular}
    }
    \label{tab:offset}
\end{table}

\vspace{0.05in}
\noindent\textbf{Combination of Region-Level Supervision and Debiased Instance Localization.} To analyze the impact of our proposed debiased instance localization (DIL), we incorporate it into the cutting-edge region-level supervision method RegionPLC~\cite{yang2023regionplc} and examine the resulting performance. As shown in Table~\ref{tab:comb_region}, the combination with DIL brings about a significant gain of 3.4\% hAP$_{50}$ on ScanNet B10/N7.
This confirms the orthogonal relationship between region-level supervision and debiased instance localization in enhancing the performance of instance localization. While region-level supervision aims to inject semantics at a finer granularity into localized 3D regions, thereby fostering a deeper understanding of 3D scenes, our debiased instance localization rectifies the objectness learning bias, ensuring more robust and generalizable proposal grouping.

\begin{table}[htbp]
    \caption{Analysis of the effectiveness of debiased instance localization (DIL) when incorporated to region-level supervision methods in terms of hAP / mAP$^\mathcal{B}$ / mAP$^\mathcal{N}$.}
    \centering
    \setlength\tabcolsep{6pt}
    \scalebox{1.0}{
        \begin{tabular}{c|c}
            \bottomrule[1pt]
            Method & \makecell{ScanNet B10/N7 \\ hAP / mAP$^\mathcal{B}$ / mAP$^\mathcal{N}$} \\ 
            \hline
            RegionPLC~\cite{yang2023regionplc} & 40.7 / \textbf{54.7} / 32.3 \\ 
            RegionPLC + DIL & \textbf{44.1} / 54.6 / \textbf{37.0} \\ 
            \toprule[0.8pt]
        \end{tabular}
    }
    \label{tab:comb_region}
\end{table}

\vspace{0.05in}
\noindent\textbf{Re-partition Experiments.} 
The robustness of our approach is further validated through a random re-sampling of base and novel categories multiple times. Specifically, we randomly re-sample the base and novel categories three times for the instance segmentation task, and we also sample the categories based on their class frequency. 
As shown in Table~\ref{tab:partitions}, Lowis3D consistently surpasses the OV-SoftGroup baseline across four different splits, achieving a substantial improvement of between 12.9\% and 55.7\% in hAP$_{50}$. This demonstrates the robustness of our approach when managing different novel classes.

\begin{table}[htbp]
    \caption{Results of experiments with re-sampled base and novel classes in terms of hAP$_{50}$ / mAP$_{50}^\mathcal{B}$ / mAP$_{50}^\mathcal{N}$.}
    \centering
    \setlength\tabcolsep{10pt}
    \scalebox{1.0}{
        \begin{tabular}{c|c|c}
            \bottomrule[1pt]
            \multirow{2}{*}{Splits}  & \multicolumn{2}{c}{hAP$_{50}$ / mAP$_{50}^\mathcal{B}$ / mAP$_{50}^\mathcal{N}$} \\
            \cline{2-3}
            & OV-SoftGroup & Lowis3D \\
            \hline
            random-sample 1 & 05.1 / 57.9 / 02.6 & \textbf{59.1} / \textbf{58.6} / \textbf{59.6} \\
            random-sample 2 & 24.4 / 53.5 / 15.8
 & \textbf{37.3} / \textbf{52.8} / \textbf{28.9} \\
            random-sample 3 & 08.9 / 55.5 / 04.8
 & \textbf{41.0} / \textbf{57.9} / \textbf{31.8} \\
            frequency-sample & 02.6 / 55.8 / 01.4
 & \textbf{58.3} / \textbf{58.1} / \textbf{58.5} \\
            \toprule[0.8pt]
        \end{tabular}
    }
    \label{tab:partitions}
\end{table}

\section{Qualitative Analysis}

\begin{figure*}[htbp]
    \begin{center}
    \includegraphics[width=1\linewidth]{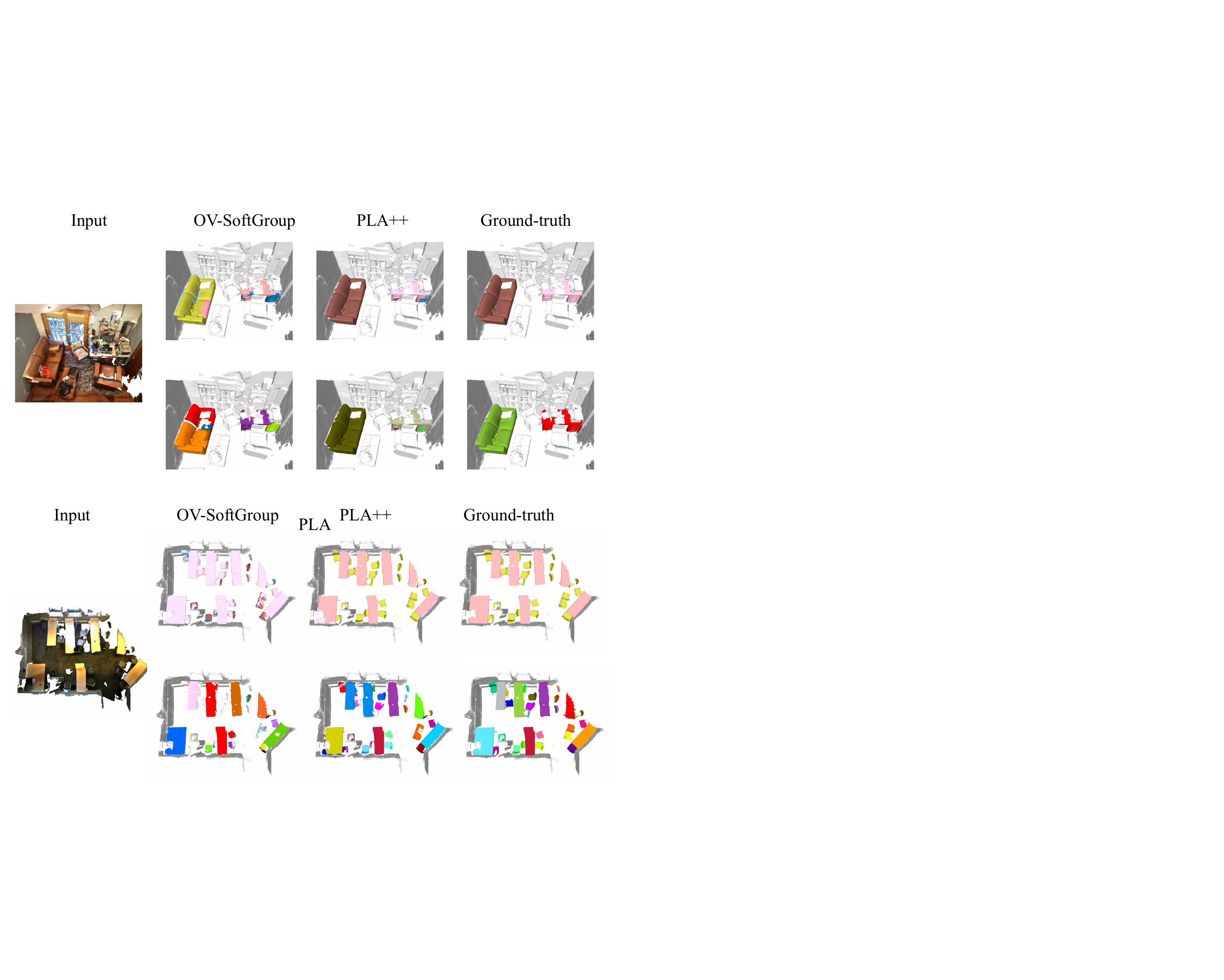}
    \end{center}
    \caption{Qualitative results of open-world instance and panoptic segmentation. Novel categories are colorized while base categories are in gray for clear differentiation. Noteworthy comparisons are highlighted within red bounding boxes.}
    \label{fig:vis_2}
\end{figure*}

To better showcase the open-world ability of our approach, we present a set of qualitative results on open-world instance segmentation and panoptic segmentation in Fig.~\ref{fig:vis_2}. In comparison to the OV-SoftGroup baseline, which frequently misclassifies unseen categories as seen categories, our Lowis3D method successfully identifies novel categories with precise semantic masks. This validates that our point-language association can inject rich semantic knowledge into the 3D encoder. Furthermore, the instance prediction masks generated by Lowis3D exhibit high accuracy, whereas OV-SoftGroup and PLA tend to either overlooks novel objects or predicts incomplete object masks. This demonstrates that our debiased instance localization greatly enhance  robustness and  generalization in localization novel categories.
Further, we present compelling qualitative results  showcasing the model's capability to recognize synonymical categories, abstract categories and even unannotated categories that are unpresent in the dataset vocabulary. 

\vspace{0.05in}
\noindent\textbf{Synonymical Novel Categories.} 
Here, we substitute class names with related yet new words during inference.
As illustrated in Fig.~\ref{fig:vis} (a), our model continues to deliver high-quality segmentation masks when we replace ``sofa'' with ``couch'' or ``refrigerator'' with ``freezer''. This demonstrates the robustness of our model in recognizing synonymous concepts.

\noindent\textbf{Abstract Novel Categories.}
Beyond object entities, our model demonstrates its capability to comprehend more abstract concepts such as types of rooms. 
As shown in Fig.~\ref{fig:vis} (b), by eliminating ``shower curtain'', ``bathtub'', ``sink'' and ``toilet'' from input categories and introducing ``bathroom'', the generated ``bathroom'' prediction generally corresponds to the actual bathroom region. 
Another example on the right illustrates the model's understanding of 'kitchen' regions. This suggests that our model is proficient in recognizing such out-of-vocabulary abstract concepts, extending beyond concrete semantic instances.

\noindent\textbf{Unannotated Novel Categories.}
Given that current 3D datasets do not annotate all classes due to prohibitive annotation costs, our model shows the potential to identity those unannotated classes with high-quality predictions, hence promoting  open-world applications.
As illustrated in Fig.~\ref{fig:vis} (c), the model successfully recognize ``monitor'' and ``blackboard'' with precise masks that are not involved in the dataset annotations. 

\section{Limitation and Open Problems}\label{sec:limitation}
While our Lowis3D framework effectively addresses open-world scene understanding by incorporating abundant semantic concepts and rectifying instance localization bias, it still faces limitations in certain areas. We highlight two main challenges here:

A key challenge is related to the performance discrepancy between S3DIS and ScanNet in open-world tasks. S3DIS demonstrates slightly lower performance attributed to its limited sample size and diversity, coupled with fewer available point-language associations. We believe that pre-training on a large dataset with rich semantic information and subsequently fine-tuning on the smaller-scale dataset or exploring dataset ensemble could be a promising alternative. This approach is left for future study and exploration.

Additionally, the calibration problem arises as the model tends to generate over-confident semantic predictions for base categories. Although we develop a binary head to calibrate semantic scores, it may face challenges in rectifying predictions for out-of-domain transfer tasks. Since the binary head is trained on dataset-specific base/novel partitions, its generalizability to other datasets with data distribution shifts is limited. This motivates us to explore and design more transferable score calibration modules in future research.

\section{Conclusion}
We propose Lowis3D, a comprehensive and efficient framework for addressing open-world instance-level 3D scene understanding. Our approach involves utilizing images as a bridge to establish hierarchical point-caption pairs, harnessing the power of 2D visual-language (VL) foundation models and the geometry relationships between 3D scenes and 2D images. Contrastive learning is employed to enhance the alignment of features in these associated pairs, thereby infusing the 3D network with a wealth of semantic concepts. Furthermore, we propose debiased instance localization to mitigate object grouping bias toward base patterns, resulting in improved generalizability in objectness learning. Extensive experiments demonstrate the effectiveness of our approach on open-world instance-level scene understanding task.


\bibliographystyle{IEEEtran}
\bibliography{IEEEabrv,egbib}

\clearpage
\appendices
\renewcommand{\thetable}{A\arabic{table}}
\renewcommand{\thefigure}{A\arabic{figure}}

\section{Dataset Category Partition}\label{sec:dataset}

\begin{table*}[htbp]
    \centering
    \caption{Category partitions for open-world instance segmentation on ScanNet. For semantic segmentation, the two background classes ``wall'' and ``floor'' are included in base categories.}
    \begin{small}
    \setlength\tabcolsep{5pt}
    \scalebox{1.0}{
        \begin{tabular}{l|l|l}
            \bottomrule[1pt]
            Partition & Base Categories & Novel Categories \\
            \hline
            B13/N4 & \makecell[l]{cabinet, bed, chair, table, door, window, picture, \\ counter, curtain, refrigerator, showercurtain, sink, bathtub} & sofa, bookshelf, desk, toilet\\
            \hline
            B10/N7 & \makecell[l]{cabinet, sofa, door, window, counter, desk, curtain, \\refrigerator, showercurtain, toilet} & bed, chair, table, bookshelf, picture, sink, bathtub \\
            \hline
            B8/N9 & \makecell[l]{cabinet, bed, chair, sofa, table, door, window, curtain} & \makecell[l]{bookshelf, picture, counter, desk, refrigerator, showercurtain, \\toilet, sink, bathtub}\\
            \toprule[0.8pt]
        \end{tabular}
    }
    \end{small}
    \label{tab:scannet_inst_partition}
\end{table*}

\begin{table*}[htbp]
    \centering
    \caption{Category partitions for open-world semantic and instance segmentation on S3DIS.}
    \begin{small}
    \setlength\tabcolsep{8pt}
    \scalebox{1.0}{
        \begin{tabular}{l|l|l}
            \bottomrule[1pt]
            Partition & Base Categories & Novel Categories \\
            \hline
            B8/N4 & \makecell[l]{ceiling, floor, wall, beam, column, door, chair, board} & window, table, sofa, bookcase \\
            \hline
            B6/N6 & \makecell[l]{ceiling, wall, beam, column, chair, bookcase} &  \makecell[l]{floor, window, door, table, sofa, board} \\
            \toprule[0.8pt]
        \end{tabular}
    }
    \end{small}
    \label{tab:s3dis_partition}
\end{table*}

\begin{table*}[htbp]
    \centering
    \caption{Category partitions for open-world panoptic segmentation on NuScenes.}
    \begin{small}
    \setlength\tabcolsep{8pt}
    \scalebox{1.0}{
        \begin{tabular}{l|l|l}
            \bottomrule[1pt]
            Partition & Base Categories & Novel Categories \\
            \hline
            B12/N3 & \makecell[l]{barrier, bicycle, bus, car, construction\_vehicle, trailer, truck,\\ driveable\_surface, sidewalk, terrain, manmade, vegetation } & motorcycle,pedestrian, traffic\_cone \\
            \hline
            B10/N5 & \makecell[l]{bicycle, bus, car, construction\_vehicle, trailer, truck, \\driveable\_surface, terrain, manmade, vegetation} &  \makecell[l]{barrier, motorcycle, pedestrian, traffic\_cone, sidewalk} \\
            \toprule[0.8pt]
        \end{tabular}
    }
    \end{small}
    \label{tab:nuscenes_partition}
\end{table*}

As mentioned in Sec. {\color{red}4.1} of the main paper, we build a 3D open-world benchmark on ScanNet~\cite{dai2017scannet}, S3DIS~\cite{armeni20163d} and nuScenes~\cite{fong2022panoptic} with multiple base/novel partitions. The concrete category partitions are shown in Table~\ref{tab:scannet_inst_partition}, Table~\ref{tab:s3dis_partition} and Table~\ref{tab:nuscenes_partition}, respectively.

\section{Usage of Images for Captioning}\label{sec:images}
For ScanNet~\cite{dai2017scannet}, we utilize a subset of 25,000 frames\footnotemark[1] from the ScanNet dataset for captioning purposes. Regarding S3DIS~\cite{armeni20163d}, due to the significant variation in the number of images per scene, we perform subsampling to ensure a maximum of 50 images per scene are used for captioning. It is worth noting that certain S3DIS scenes do not have corresponding images available, which means we cannot provide language supervision for those scenes during the training process. Lastly, for nuScenes~\cite{fong2022panoptic}, we utilize all available images in the dataset.

\footnotetext[1]{\url{https://kaldir.vc.in.tum.de/scannet_benchmark/documentation}}

\section{Caption Examples}\label{sec:caption_exps}

\begin{figure*}[htbp]
    \begin{center}
    \includegraphics[width=1\linewidth]{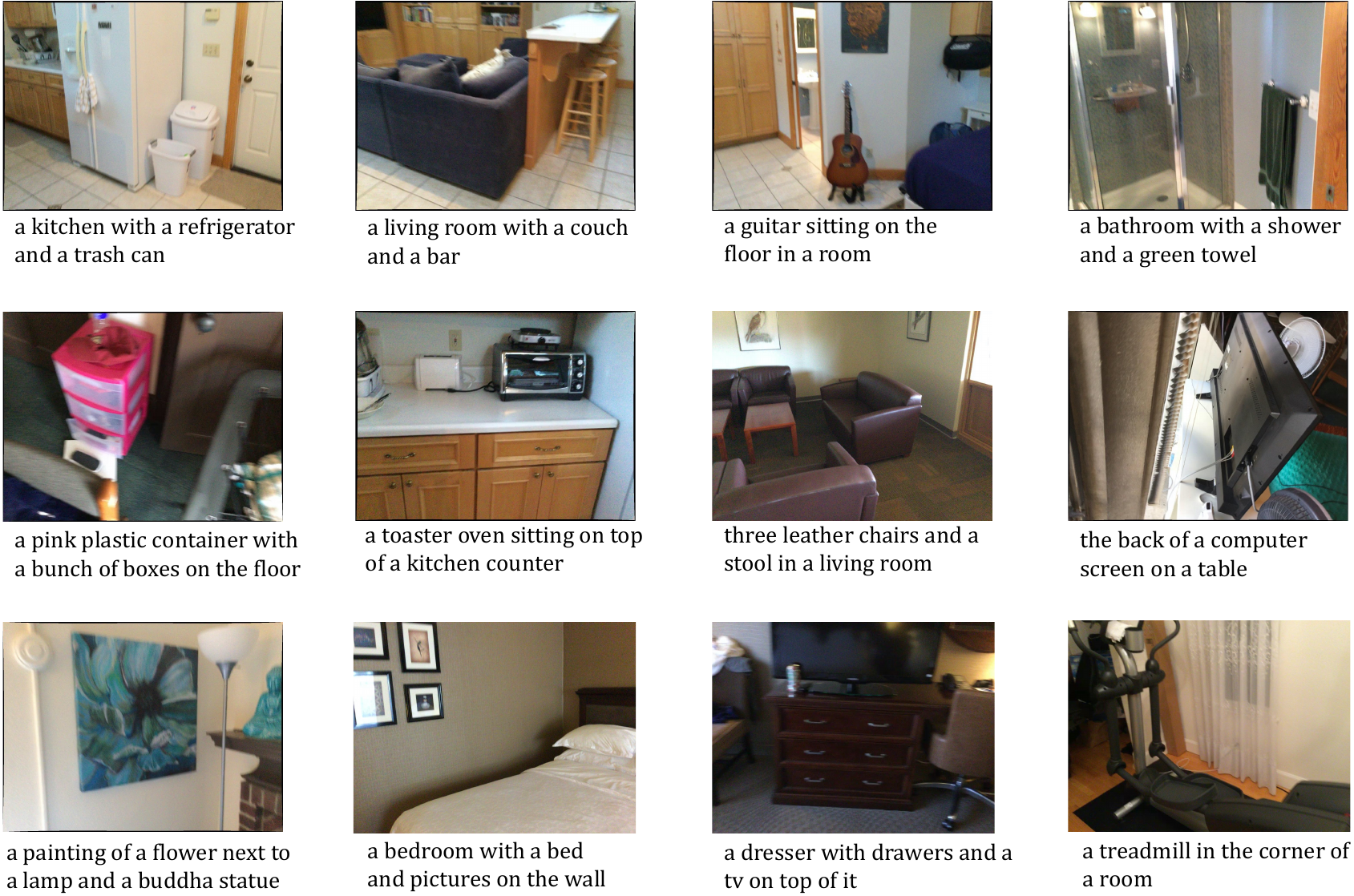}
    \end{center}
    \caption{Examples of image-caption pairs by image-captioning model ViT-GPT2~\cite{vit-gpt2}.}
    \label{fig:image-caption}
\end{figure*}

\begin{figure*}[htbp]
    \begin{center}
    \includegraphics[width=1\linewidth]{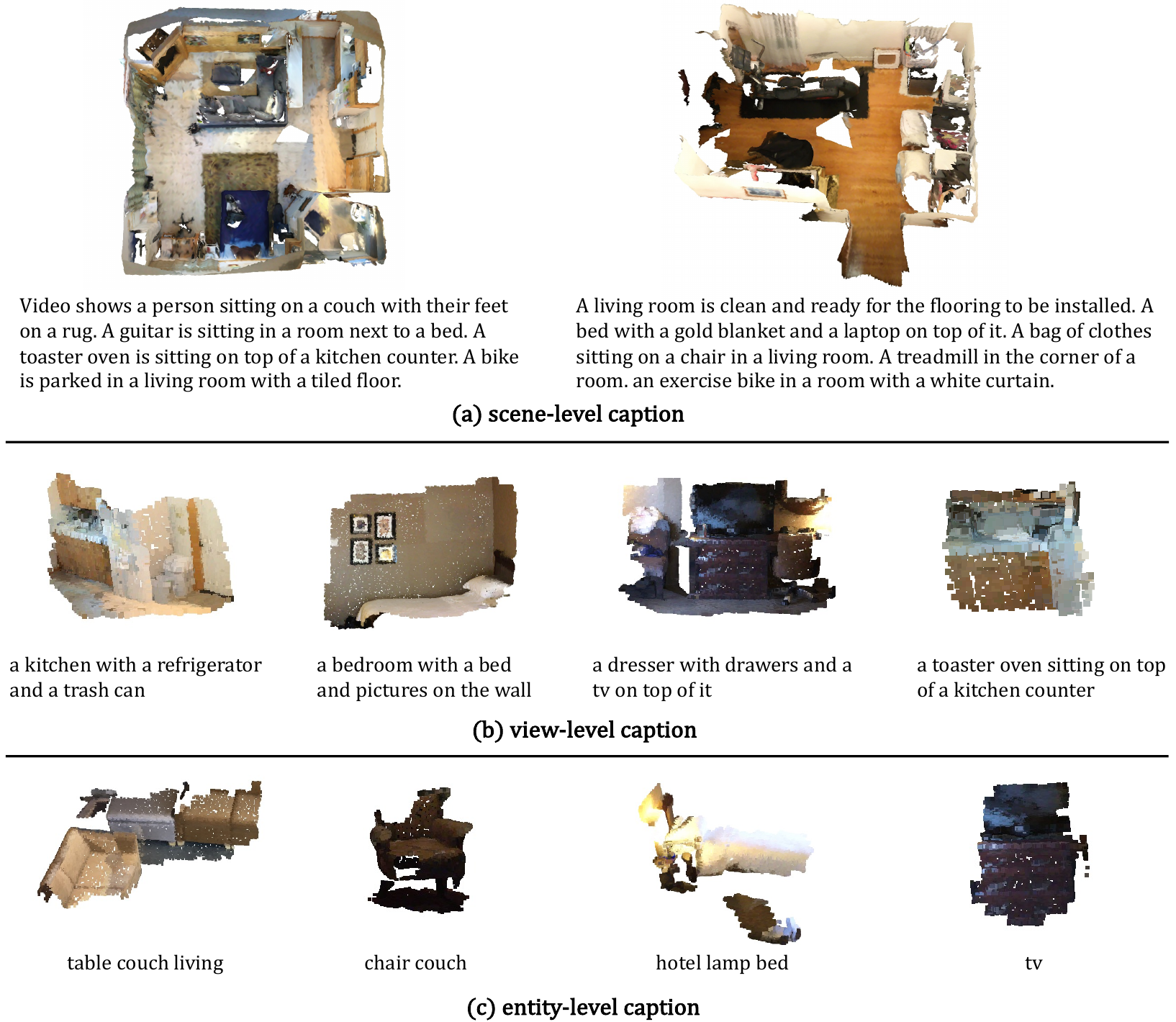}
    \end{center}
    \caption{Examples of hierarchical point-caption pairs from ScanNet~\cite{dai2017scannet}.}
    \label{fig:point-caption}
\end{figure*}

In this section, we provide examples of image-caption pairs generated by vision-language (VL) foundation models, as well as examples of hierarchical associated point-caption pairs.

As depicted in Fig.~\ref{fig:image-caption}, the image captions effectively describe the main entities present in the images, along with room types (\eg kitchen), textures (\eg leather), colors (\eg green) and spatial relationships (\eg on top of). These captions convey rich semantic clues with a large vocabulary size. Notably, even uncommon classes such as ``buddha statue'' are correctly detected, highlighting the generalizability of existing VL foundation models and the semantic comprehensiveness of the generated captions.

With the obtained image-caption pairs, we can hierarchically associate 3D points and captions by leveraging geometric constraints between 3D point clouds and multi-view images.
As shown in Fig.~\ref{fig:point-caption} (a), the scene-level caption describes each area/room (\eg kitchen, living room) in the in the entire scene, providing abundant vocabulary and semantic-rich language supervision. The view-level caption in Fig.~\ref{fig:point-caption} (b) focuses on single view frustums of the 3D point cloud, capturing more local details with elaborate text descriptions.  This enables the model to learn region-wise vision-semantic relationships.  Additionally, as shown in Fig.~\ref{fig:point-caption} (c), the entity-level caption covers only a few entities within small 3D point sets with concrete words as captions, providing more fine-grained supervisions to facilitate learning of object-level understanding and localization.

\end{document}